\documentclass[letterpaper, 10 pt, conference]{ieeeconf}  

\IEEEoverridecommandlockouts                              

\usepackage[utf8]{inputenc}
\usepackage{epsfig} 
\usepackage{mathptmx} 
\usepackage{times} 
\usepackage{amsmath} 
\usepackage{amssymb}  

\usepackage{array}

\title{\LARGE \bf
AquaFeat+: an Underwater Vision Learning-based Enhancement Method for Object Detection, Classification, and Tracking
}

\author{Emanuel C. Silva$^{1}$, Tatiana T. Schein$^{1}$, José D. G. Ramos$^{1}$, Eduardo L. Silva$^{1}$,\\Stephanie L. Brião$^{1}$, Felipe G. Oliveira$^{2}$ and Paulo L. J. Drews-Jr$^{1}$
\thanks{*The authors acknowledge the financial support of CNPq, FINEP, FAURG, and FAPERGS. This research was supported by the Human Resources Program of the National Agency of Petroleum, Natural Gas, and Biofuels (PRH/ANP–PRH22.1/FURG). We also acknowledge the support of the São Paulo Research Foundation (FAPESP), Brazil, under grant number 2024/10523-5.}
\thanks{$^{1}$Centro de Ciências Computacionais (C3), Universidade Federal do Rio Grande (FURG), Brazil. {\tt\footnotesize\{emanuel\_silva,tatischein,jgarcia,eduardolawson,
stephanie.loi,paulodrews\}
@furg.br}}
\thanks{$^{2}$Instituto de Ciências Exatas e Tecnologia (ICET), Universidade Federal do Amazonas (UFAM), Brazil.  {\tt\footnotesize felipeoliveira@ufam.edu.br}}%
}

\begin{document}

\maketitle
\thispagestyle{empty}
\pagestyle{empty}

\begin{abstract}

Underwater video analysis is particularly challenging due to factors such as low lighting, color distortion, and turbidity, which compromise visual data quality and directly impact the performance of perception modules in robotic applications. This work proposes AquaFeat+, a plug-and-play pipeline designed to enhance features specifically for automated vision tasks, rather than for human perceptual quality. The architecture includes modules for color correction, hierarchical feature enhancement, and an adaptive residual output, which are trained end-to-end and guided directly by the loss function of the final application. Trained and evaluated in the FishTrack23 dataset, AquaFeat+ achieves significant improvements in object detection, classification, and tracking metrics, validating its effectiveness for enhancing perception tasks in underwater robotic applications.

\end{abstract}

\section{INTRODUCTION}

Autonomous and remotely operated underwater robots play a major role in the exploration and monitoring of marine environments, supporting tasks in marine biology, environmental science, and natural resource management. In these robotic applications, computer vision systems, particularly object detection in video, enable critical operations such as species identification and tracking, debris monitoring, and infrastructure inspection. However, the underwater environment poses severe challenges to vision algorithms, mainly due to adverse conditions \cite{schein2025udbe}. Videos captured in underwater scenarios suffer from color distortions, light attenuation, and scattering effects that cause loss of contrast and sharpness. These degradations not only limit target visibility but also drastically compromise the performance of object detection, classification, and tracking models \cite{dawkins2024fishtrack23}.

Underwater robotic systems rely heavily on high-quality visual data to perform the aforementioned tasks. However, adverse underwater conditions often compromise image quality, making visual enhancement models a necessary tool for improving video usability. While these approaches, particularly those based on deep learning, can significantly enhance image appearance, they often come with high computational demands and long processing times, which affect their applicability in real-time robotic operations. Furthermore, most existing enhancement methods are designed to optimize perceptual quality for human observers, rather than maximizing the preservation or extraction of features critical for automated analysis. Consequently, an image that looks visually appealing may still obscure or distort essential patterns and textures, negatively impacting the performance of object detection, classification, and tracking algorithms \cite{srinath2025undive}. These limitations highlight the need for specialized enhancement strategies that are both computationally efficient and designed to support downstream vision tasks in underwater robotics.



In this work, we propose AquaFeat+, an innovative and efficient module that enhances underwater videos directly at the feature level. Designed as a plug-and-play component, it seamlessly integrates with existing architectures to optimize object detection, classification, and tracking. We evaluated its effectiveness across different tasks by combining it with multiple SOTA architectures, using the YOLOv8m for detection/tracking and the YOLOv11s-cls backbone for classification, in relation to the FishTrack23 dataset \cite{dawkins2024fishtrack23}. The performance of our method was benchmarked against its predecessor (AquaFeat \cite{Silva2025AquaFeat_unpub}), other enhancement techniques (FeatEnHancer \cite{hashmi2023featenhancer}), and prominent models in each respective task, such as YOLOv8s \cite{Hasib_Underwater_Object_Detection_2024_commit}, and ConvNeXt \cite{liu2022convnet}. When evaluated across key underwater perception tasks, AquaFeat+ demonstrates highly competitive performance. It achieves the highest F1-Score in object detection, delivers leading classification results with top scores for precision, recall, and F1-Score, and excels in multi-object tracking by obtaining the highest HOTA score. These results validate its ability to significantly elevate perception capabilities in complex underwater scenes. 

The main contributions of this work are:

\begin{itemize}
    \item We propose AquaFeat+, a plug-and-play module designed to enhance hierarchical features in low-light underwater videos, optimizing performance for object detection, classification, and tracking tasks.
    \item We demonstrate the effectiveness of training AquaFeat+ end-to-end, guided directly by the loss function of the final task, ensuring that enhancement is directly relevant to the computer vision application.
    \item We adapt the FishTrack23 \cite{dawkins2024fishtrack23} dataset with a focus on optimizing object detection, classification, and tracking tasks for robotic perception.
\end{itemize}

\section{Related Work}

Computational analysis of underwater environments presents unique challenges due to image degradation caused by light absorption and scattering within the water column. This section highlights advances in pertinent areas to our proposal, such as underwater image and video enhancement. Furthermore, task analysis includes object detection, classification, and tracking.


\subsection{Underwater Image and Video Enhancement}

Underwater image and video enhancement methods can be categorized as visually or feature-based. Visual-based approaches improve the aesthetic and perceptual quality of images, while feature-based approaches enhance specific attributes for particular tasks. Supervised methods train neural networks to map degraded underwater data to clear references, often using synthetic data. These deep neural networks are designed to address low-light conditions, enhance colorization, and minimize glare \cite{hashmi2023featenhancer} \cite{colorcorrectionANDSpecialConv}. For videos, they can also leverage temporal redundancy across frames to boost performance.

In contrast, unsupervised methods do not require paired data. These include Generative Adversarial Networks (GANs), which use lightweight networks trained with reference-free losses, and diffusion models. For underwater videos, a probabilistic denoising diffusion model is used to learn a generative prior \cite{srinath2025undive}. Multiframe-to-multiframe networks \cite{li2025multiframe} enhance underwater videos without relying on pairwise coding, which is particularly relevant in the underwater domain, where data collection is challenging.

Recent works propose feature-based enhancement methods specifically designed to improve downstream tasks. The FeatEnHancer model \cite{hashmi2023featenhancer}, although developed for terrestrial environments, uses task-centric end-to-end learning to optimize features for the final metric rather than purely visual quality, making it suitable for underwater domains. Following this principle, the AquaFeat model \cite{Silva2025AquaFeat_unpub} introduces a plug-and-play feature enhancement module that integrates directly with YOLO-based detectors. By performing multi-scale feature aggregation and task-guided optimization, AquaFeat improves detection precision and recall while maintaining computational efficiency. Other models, such as AMSP-UOD \cite{zhou2024amsp} and Multi-Scale Feature Enhancement \cite{li2025multi}, combine image formation priors and neural optimization to mitigate degradation and preserve fine-grained details. Despite these advances, image enhancement and feature-based tasks are still often handled in independent stages, reinforcing the need for task-oriented methodologies. Our AquaFeat+ directly addresses this gap by integrating feature enhancement directly into the vision task pipeline, optimizing for downstream performance rather than just visual appeal, and adapting these concepts specifically for the challenging underwater environment.

\subsection{Underwater Video Analysis for Vision Tasks}

Underwater video analysis for tasks like object detection, classification, and tracking is a challenging field where deep learning (DL) models are increasingly applied to overcome visual degradation.

\textbf{Object detection} in underwater environments faces unique challenges such as low visibility, color distortion, and the presence of small or occluded objects. Real-time models like YOLO are widely adopted, with recent advancements focusing on enhancing feature extraction and robustness to these conditions, often through improved architectures or specialized modules \cite{chen2024dynamic}. Performance is boosted either by pre-processing the images or by integrating feature enhancement directly into the detection pipeline, as seen with AquaFeat \cite{Silva2025AquaFeat_unpub}.

\textbf{Object classification}, particularly for species recognition, benefits from advanced architectures like ConvNeXt V2 \cite{liu2022convnet}. This task is advanced by works like WildFish \cite{zhuang2018wildfish}, which introduced new challenges such as open-set classification and multi-modal, text-guided fine-grained recognition to handle unseen or highly similar species. Performance is further boosted by enhancements at the feature level.

\textbf{Object tracking} typically follows the tracking-by-detection paradigm, where associating detections across frames is the primary challenge in poor visibility and frequent occlusions. Methods like ByteTrack \cite{zhang2022bytetrack} are well-suited for these conditions by leveraging low-confidence detections to maintain object trajectories. Recent efforts also explore multi-object tracking (MOT) in complex underwater scenarios, often incorporating memory aggregation or robust association strategies to handle long-term tracking and re-identification \cite{dawkins2024fishtrack23}.

\section{AquaFeat+ Model}

\begin{figure*}
\centering
\includegraphics[width=\textwidth]{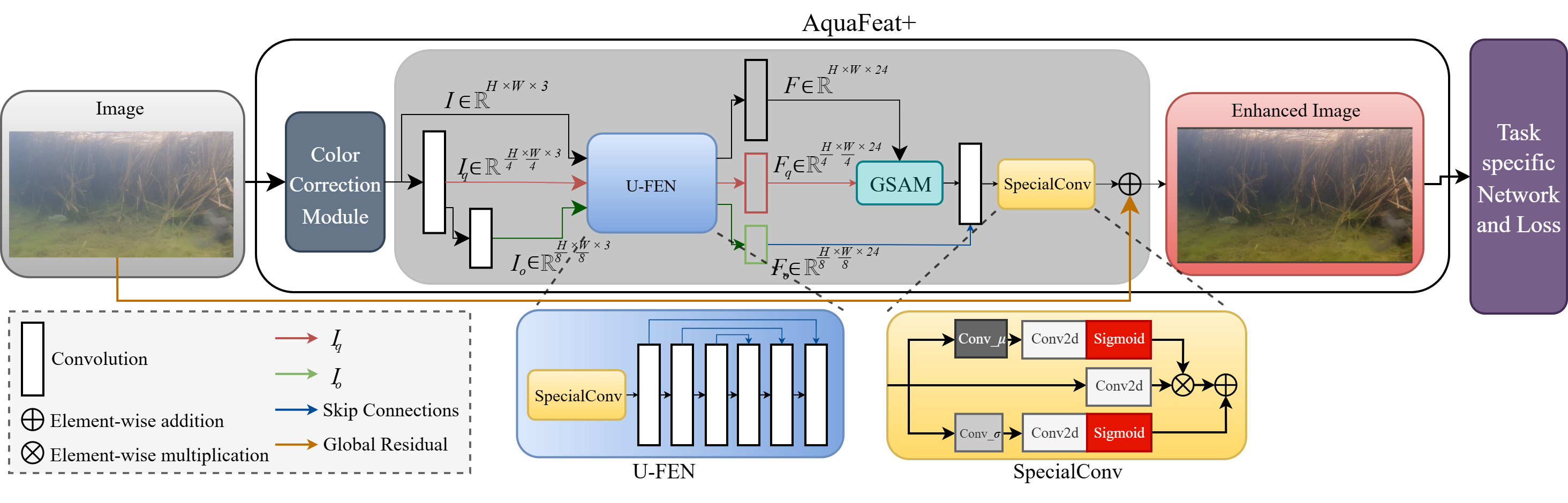}
\caption{Overview of the AquaFeat+ architecture. The pipeline begins with the color correction module, followed by the Underwater-Feature Enhancement Network (U-FEN) for feature extraction. The Global-Scale Attention Module (GSAM) aggregates global features and merges them with those from the original 1/8th-sized image to produce an enhanced image, which is then fed into the chosen task backbone.}
\label{arquitetura}
\end{figure*}

Inspired by the original AquaFeat method \cite{Silva2025AquaFeat_unpub}, AquaFeat+ is a plug-and-play video enhancement model designed to address challenges in underwater video analysis, such as low light and color distortions. Its primary goal is to provide a higher-quality, feature-rich input signal to improve the performance of downstream computer vision tasks like object detection, classification, and tracking. The architecture comprises three main modules: Color Correction, Feature Enhancement, and an Adaptive Residual Output, as illustrated in Fig. \ref{arquitetura}.

\begin{figure}
\centering 
\includegraphics[width=0.75\columnwidth]{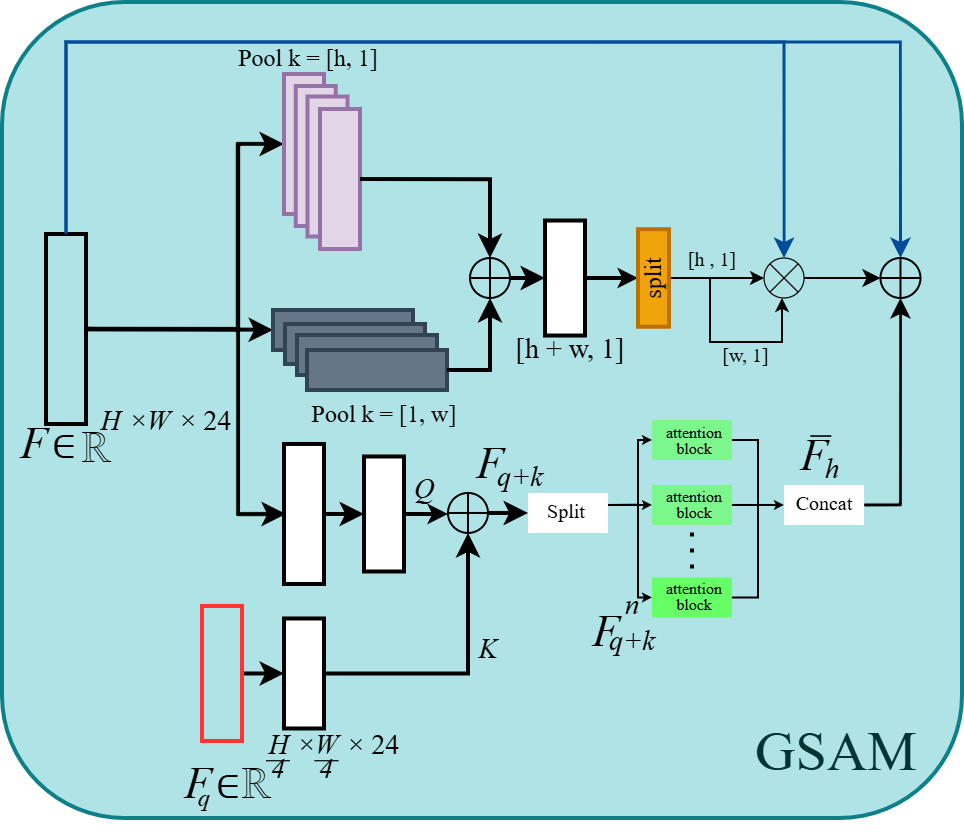}
\caption{An overview of the Global-Scale Attention Module (GSAM), designed to enhance features by incorporating both global and multi-scale context. The module first processes the input along two distinct paths. One path captures long-range spatial dependencies using a global attention mechanism. The second path intelligently fuses features from different scales. The outputs of paths are combined with the original input feature map to produce a comprehensive and enriched representation.}
\label{fig:gsam}
\end{figure}


\subsection{Color Correction Module}

The underwater environment has challenges to visual quality, such as chromatic distortions and selective absorption of light wavelengths. To mitigate these consequences, AquaFeat+ incorporates a color correction module that acts as a non-trainable preprocessing step. This module analyzes the average intensities of the color channels (red, green, and blue) and adjusts the color distribution to the median channel \cite{colorcorrectionANDSpecialConv}. This initial white balance equalization can reduce the severe visual degradations in underwater videos, providing cleaner, feature-rich input for subsequent vision models. The main goal is to ensure that subsequent layers focus on extracting more complex features without being hampered by color artifacts.


\subsection{Feature Enhancement Process}

Adapted from the AquaFeat method \cite{Silva2025AquaFeat_unpub}, AquaFeat+'s feature enhancement process seamlessly integrates with downstream vision tasks, enhancing hierarchical features at multiple levels. This process is subdivided into three stages as described below.


\subsubsection{Underwater-Feature Enhancement Network} After color correction, the RGB images of the video frames are processed in three parallel streams, at their original, one-quarter, and one-eighth resolutions. Each stream is fed into an Underwater Feature Enhancement Network (U-FEN) with shared weights, inspired by image enhancement methods for harsh conditions. The U-Net network replaces the conventional first layer with a SpecialConv \cite{colorcorrectionANDSpecialConv} layer. Unlike a standard convolution, SpecialConv introduces a trainable, content-aware mechanism that dynamically adjusts contrast. Calculate per-channel statistics (mean and standard deviation) from its input to generate adaptive multipliers, which are applied to the convolutional features to enhance their distinctiveness. After this layer, the activation function, LeakyReLU, is applied, which helps prevent the problem of vanishing gradients during feature processing. The rest of the U-FEN encoder uses six standard convolutional layers ($3 \times 3$ kernel) with dense skip connections to effectively process and enhance features at each scale.


\subsubsection{Global-Scale Attention Module} 

The Global-Scale Attention Module (GSAM) is a hybrid neural network architecture engineered to enhance feature representations through a dual-pathway design (see Fig. \ref{fig:gsam}). It concurrently performs intra-scale feature refinement and inter-scale feature fusion. The module's architecture is founded on two parallel sub-modules: a Global Feature-Aware (GFA) \cite{zhou2024amsp} module for spatial enhancement and an enhanced Scale-Aware Feature Aggregation (SAFA) \cite{hashmi2023featenhancer} mechanism for multi-scale fusion. These components, operating in parallel, are integrated with the original input via a residual connection, forming a composite block that enriches features with both spatial and scale-based contexts without losing foundational information.

The two core sub-modules address distinct aspects of feature enhancement. The GFA module operates exclusively on the primary feature map to explicitly model long-range spatial dependencies. It achieves this by generating coordinate-informed attention maps that modulate the input feature, selectively amplifying salient spatial regions. In parallel, the SAFA component performs inter-scale fusion. It takes both the high-resolution feature map and a lower-resolution context map, projects them into a common embedding space, and computes attention weights to perform a weighted aggregation. This process intelligently blends information from different scales, capturing a more holistic representation.

The final output of the GSAM is the element-wise summation of three distinct tensors: the original input feature map (preserved via a skip connection), the spatially-refined feature map from the GFA module, and the aggregated multi-scale feature map from the SAFA mechanism. This tripartite summation is the key to the module's efficacy. By integrating a residual path and the outputs of two specialized, parallel pathways, GSAM produces a final feature map that is simultaneously enhanced with global spatial context and rich multi-scale information, leading to a more robust and expressive representation for downstream tasks.

\subsubsection{Final Feature Aggregation}

The output of the GSAM module is integrated with the features of the path with the lowest resolution (one-eighth) to complete feature fusion. Firstly, the small-scale feature map is resized via bilinear interpolation to match the spatial dimensions of the GSAM output. These two feature streams are then concatenated along the channel dimension, creating a unified tensor that combines broad contextual information with fine-grained details. This tensor is processed by a final convolutional layer $3\times3$, which learns to merge the features into a single comprehensive representation before being passed to the adaptive output stage.

\subsection{Adaptive Residual Output}


The final stage of AquaFeat+ synthesizes enhanced images through a terminal SpecialConv layer, which generates an enhancement residual map. This residual, normalized using a tanh activation for stability, is added to the original input image, preserving structural integrity while applying learned enhancements. The model is trained using task-specific loss functions from downstream applications (object detection, classification, tracking), ensuring the enhancement is directly aligned with the target task requirements rather than generic image quality metrics.

\section{EXPERIMENTAL RESULTS}

\subsection{Dataset}

To adapt the FishTrack23 \cite{dawkins2024fishtrack23} dataset for the proposed tasks, a preprocessing script was developed. The initial stage consisted of data cleaning, where videos without annotations and duplicate annotations were discarded. Additionally, a manual inspection was performed to exclude videos in which the annotated objects did not correspond to fish, such as plants and lures. Subsequently, a frame extraction strategy was implemented. For the training and validation sets, one frame was extracted for every $20$ frames containing fish annotations. For the test set, which was already partitioned, all frames with annotations were extracted, supplemented by $5\%$ of frames from the video without annotations to allow video reconstruction during inference. The dataset resulting from this stage totaled $5,149$ images for training, 1,098 for validation, and $14,575$ images (between $23$ and $46$ minutes of diverse videos, since some were annotated at a 5Hz rate and others at 10Hz) for testing.

This dataset served as the basis for a second processing stage, specific to the classification methods. In this phase, the images were cropped based on the bounding boxes to isolate only the fish. To mitigate the severe class imbalance present in the original dataset of $73$ species, a grouping of classes was performed: the three most representative classes, \textit{Lutjanus campechanus}, \textit{Micropterus salmoides}, and \textit{Pagrus pagrus}, which together constituted over $50\%$ of the samples, were retained. All remaining $70$ classes were grouped into a single category named "unspecified fish." Following this categorization, another manual verification was conducted to remove images where the fish was not clearly visible or the annotation was wrong. Finally, the resulting dataset was split into training ($80\%$) and validation ($20\%$). For testing, we used the same images from the test detection set, cropped the fish, and applied the same idea.

\subsection{Qualitative Evaluation}

\begin{figure*}
    \centering
    
    \setlength{\tabcolsep}{0pt}
    
    \scalebox{0.9}{
    \begin{tabular}{
        >{\centering\arraybackslash}m{2.2cm} 
        >{\centering\arraybackslash}m{0.15\textwidth} 
        @{} 
        >{\centering\arraybackslash}m{0.15\textwidth}
        @{} 
        >{\centering\arraybackslash}m{0.15\textwidth}
        >{\centering\arraybackslash}m{0.8cm} 
        >{\centering\arraybackslash}m{0.15\textwidth} 
        @{}
        >{\centering\arraybackslash}m{0.15\textwidth}
        @{}
        >{\centering\arraybackslash}m{0.15\textwidth}
    }
    \scriptsize Ground Truth &
    \includegraphics[width=\linewidth]{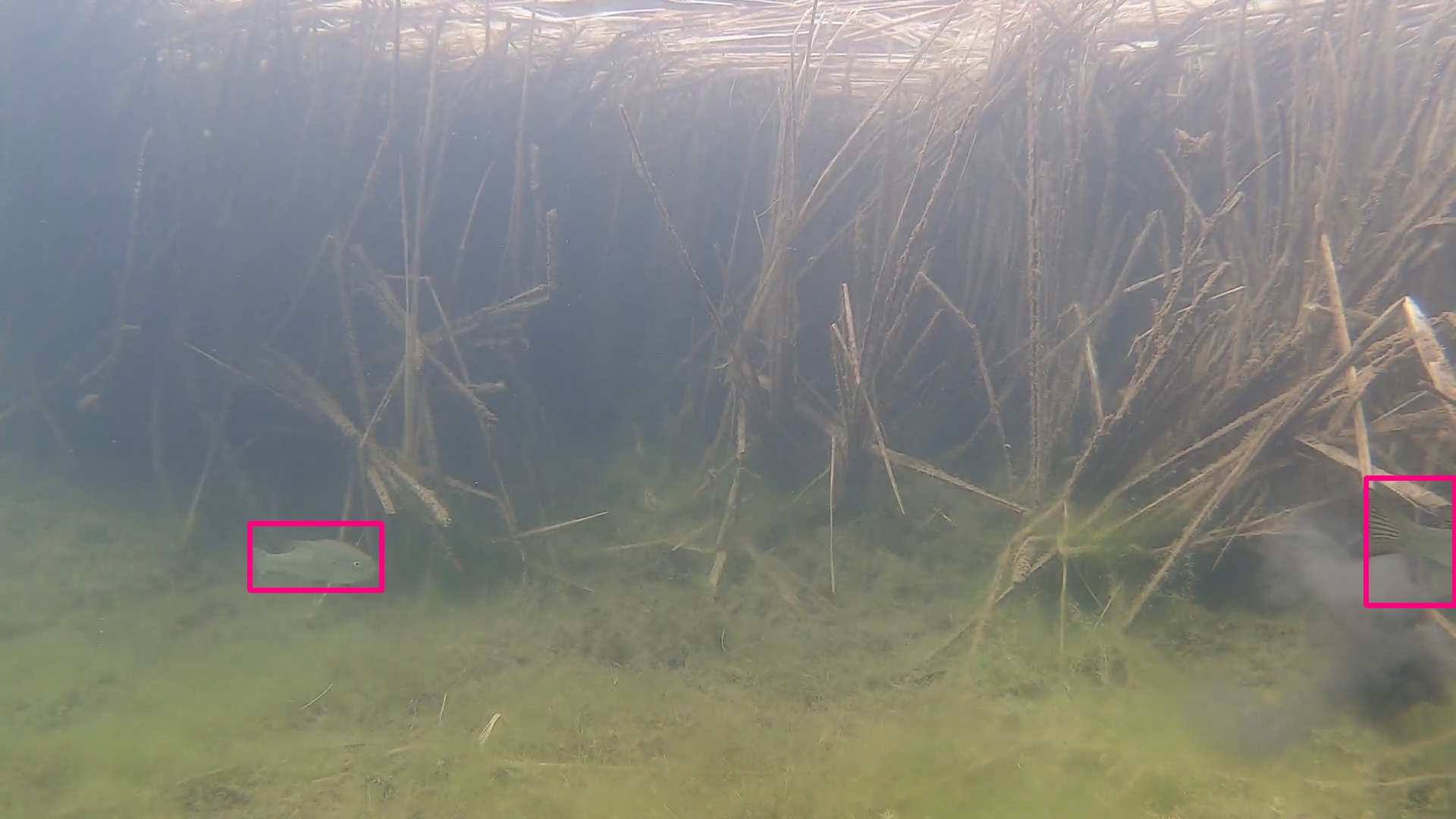} &
    \includegraphics[width=\linewidth]{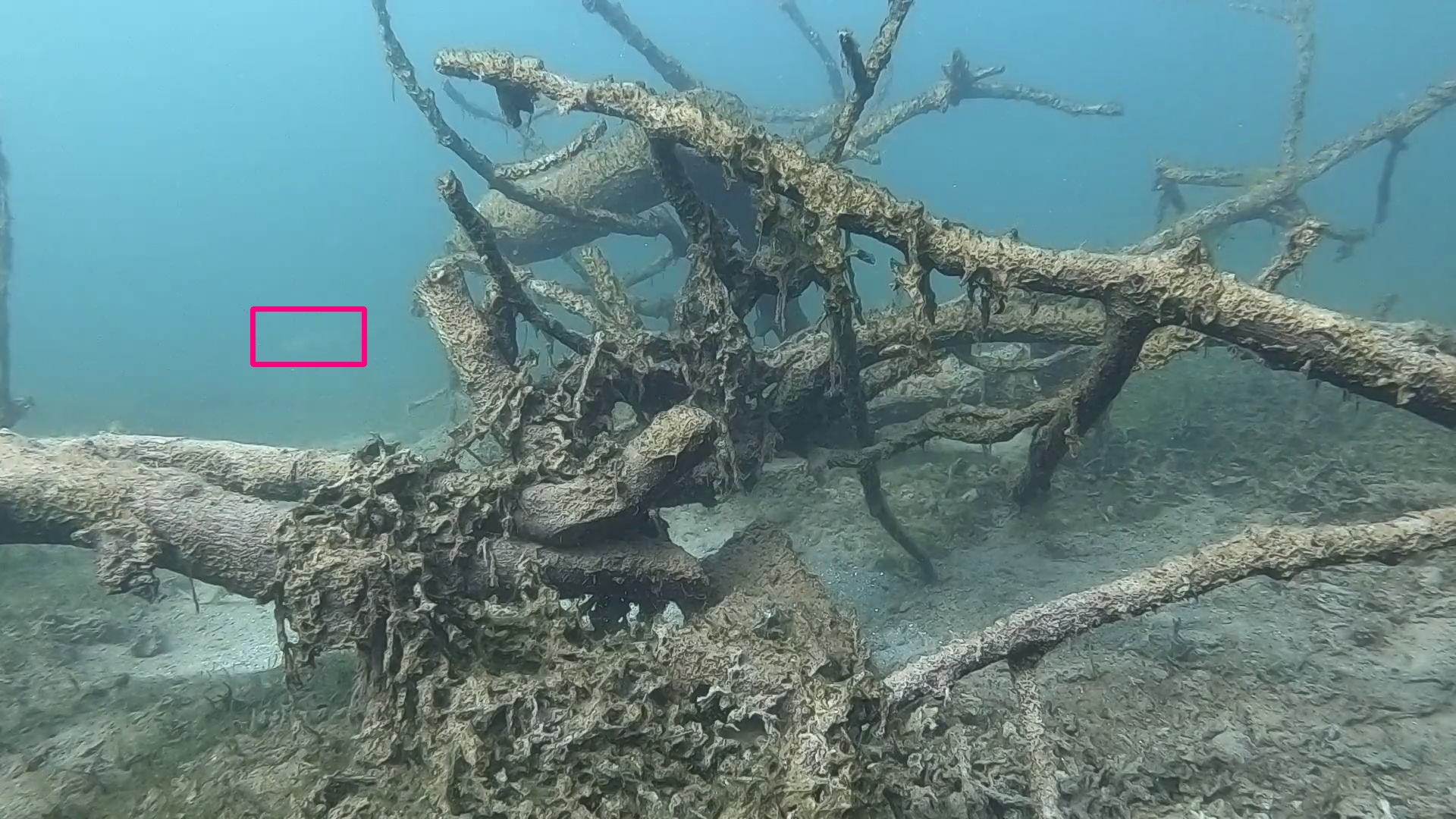} &
    \includegraphics[width=\linewidth]{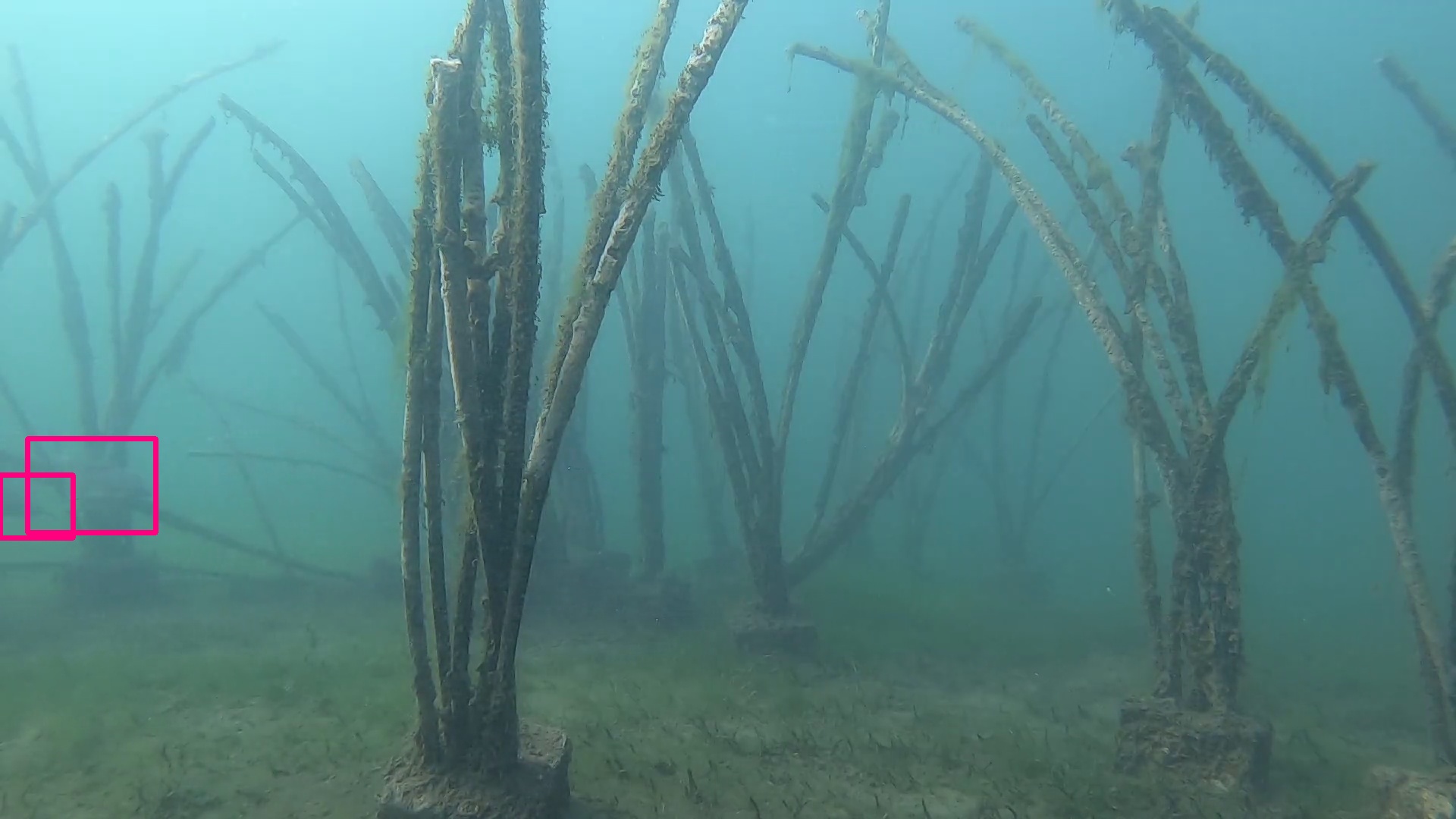} &
    & 
    \includegraphics[width=\linewidth]{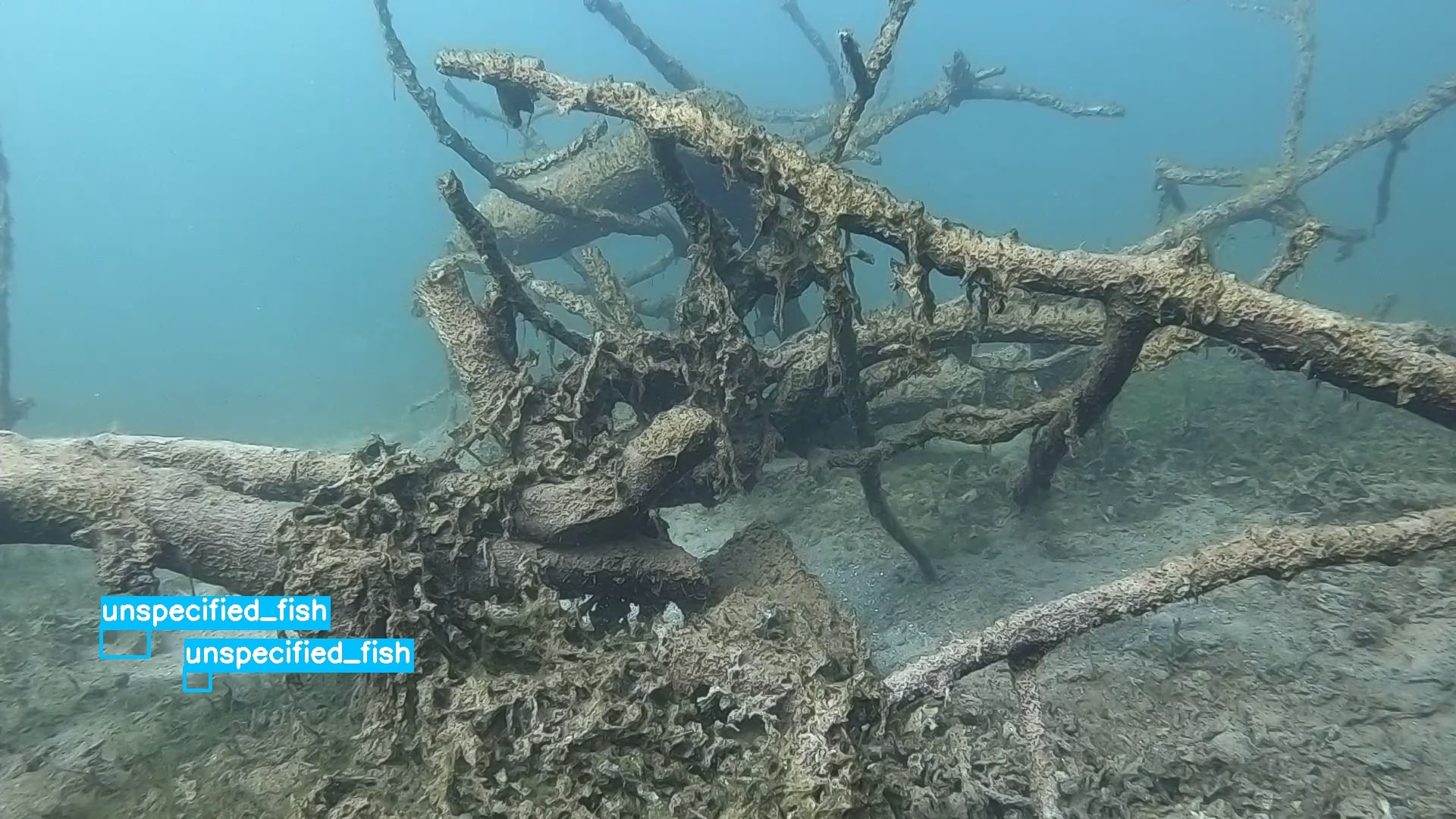} &
    \includegraphics[width=\linewidth]{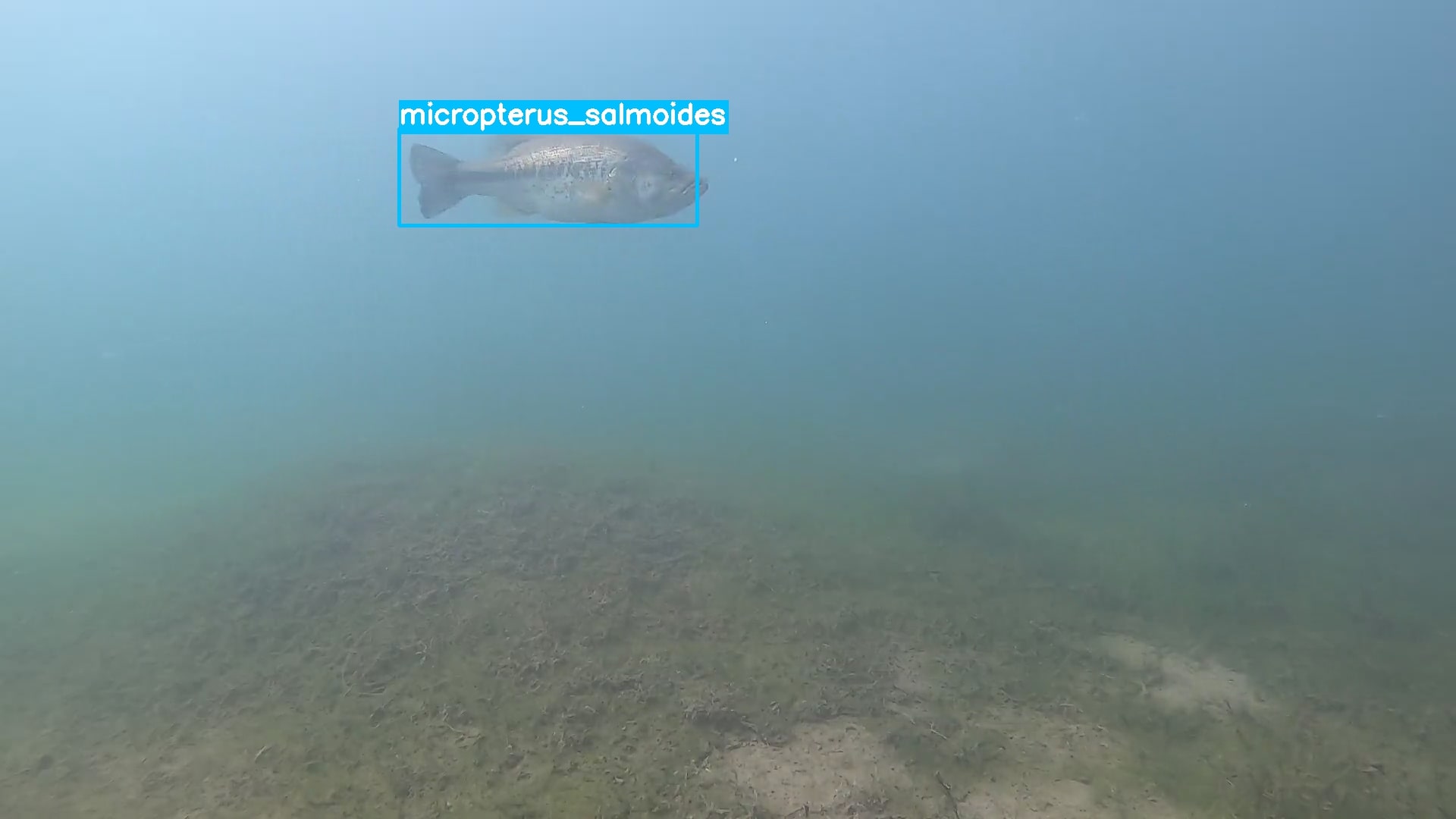} &
    \includegraphics[width=\linewidth]{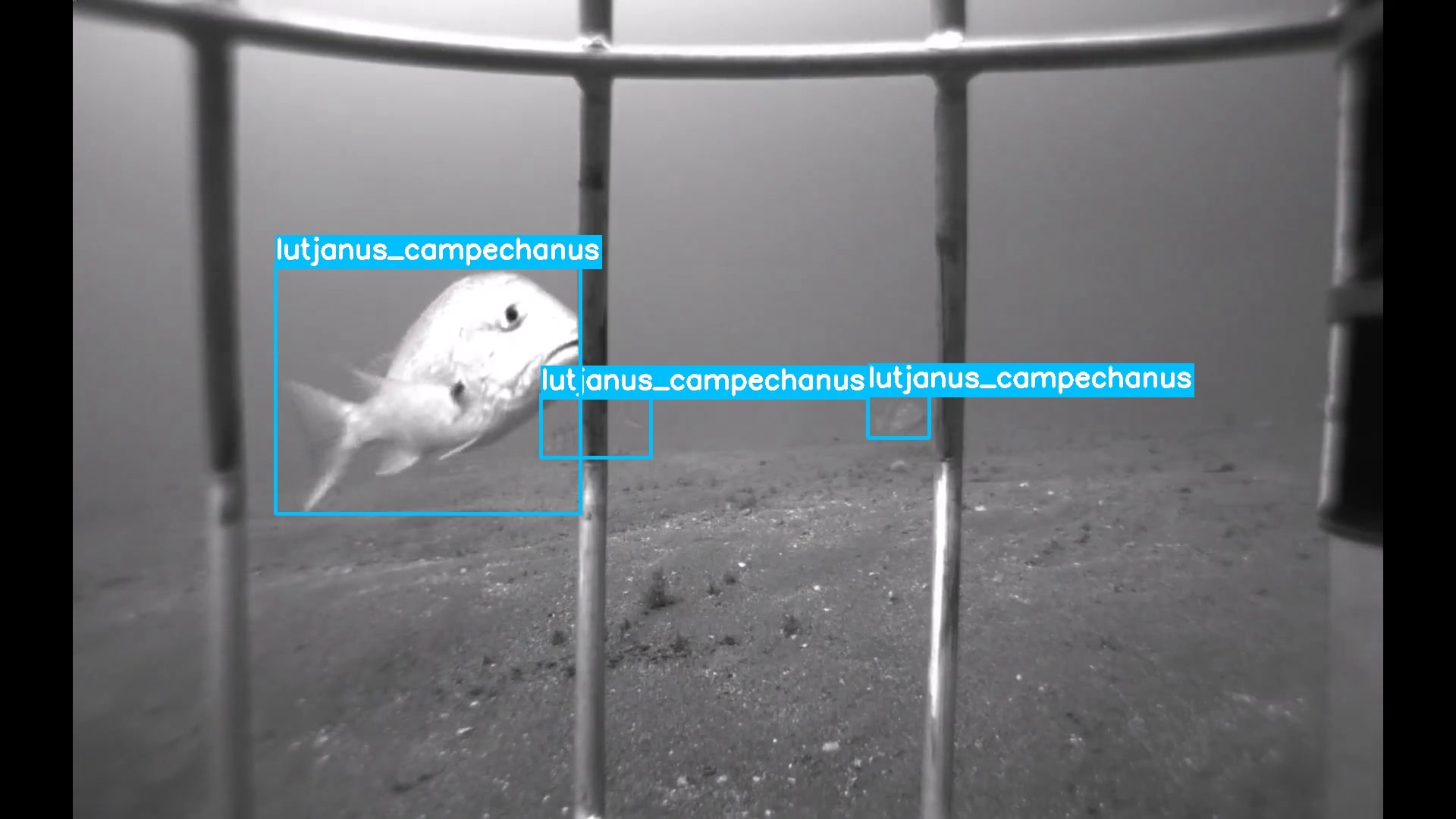} \\[-4pt] 
    
    \scriptsize YOLO &
    \includegraphics[width=\linewidth]{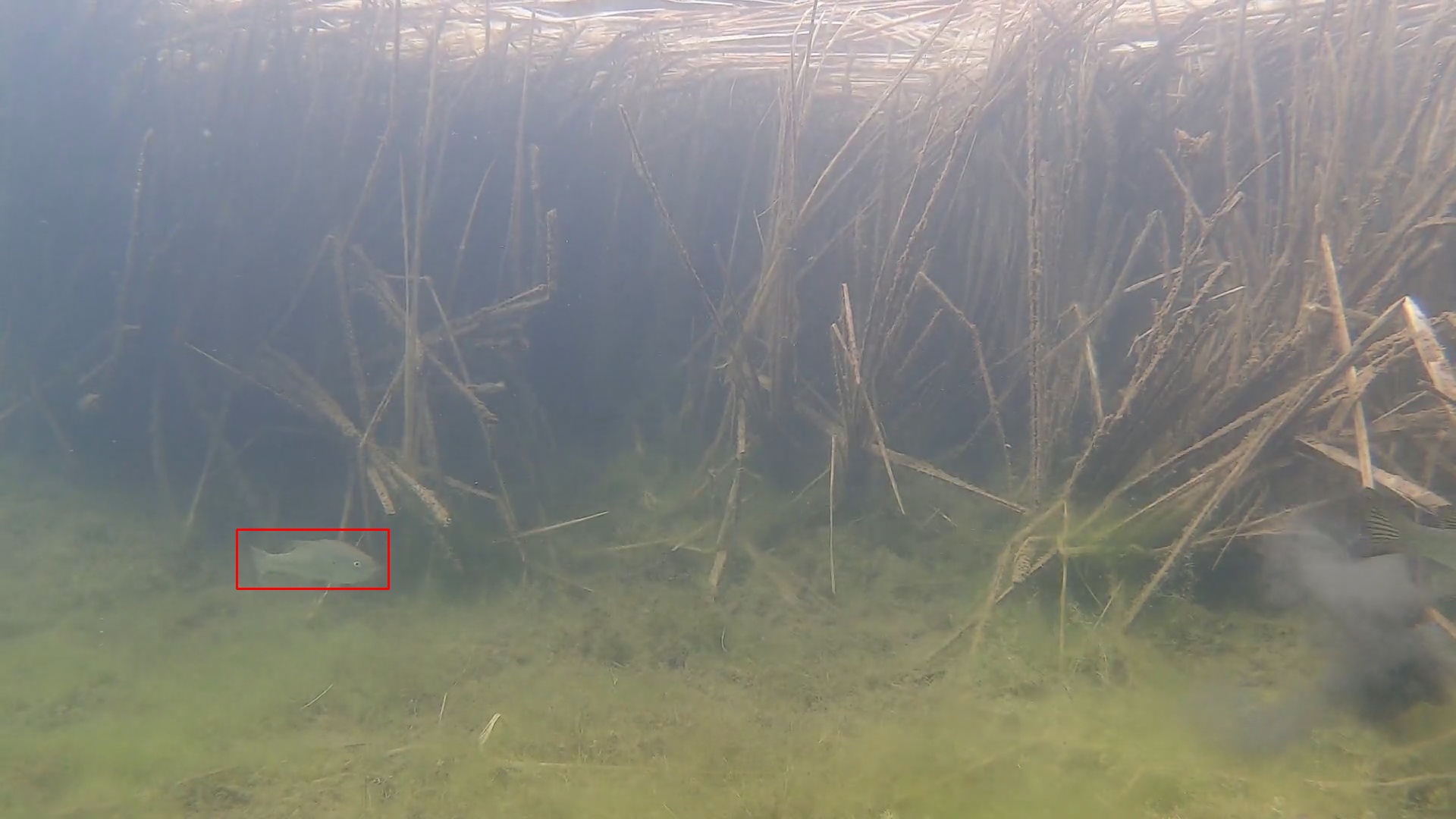} &
    \includegraphics[width=\linewidth]{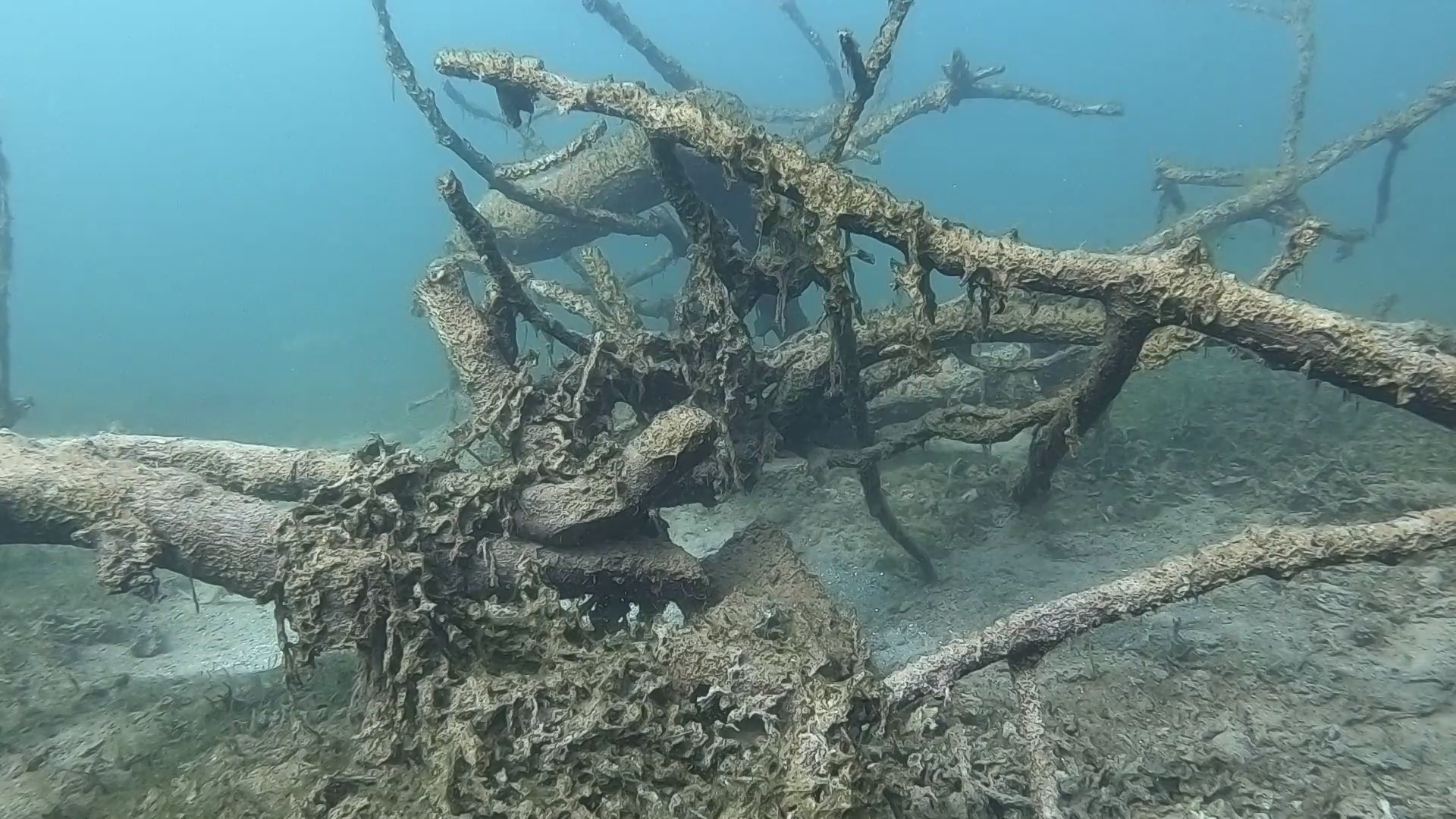} &
    \includegraphics[width=\linewidth]{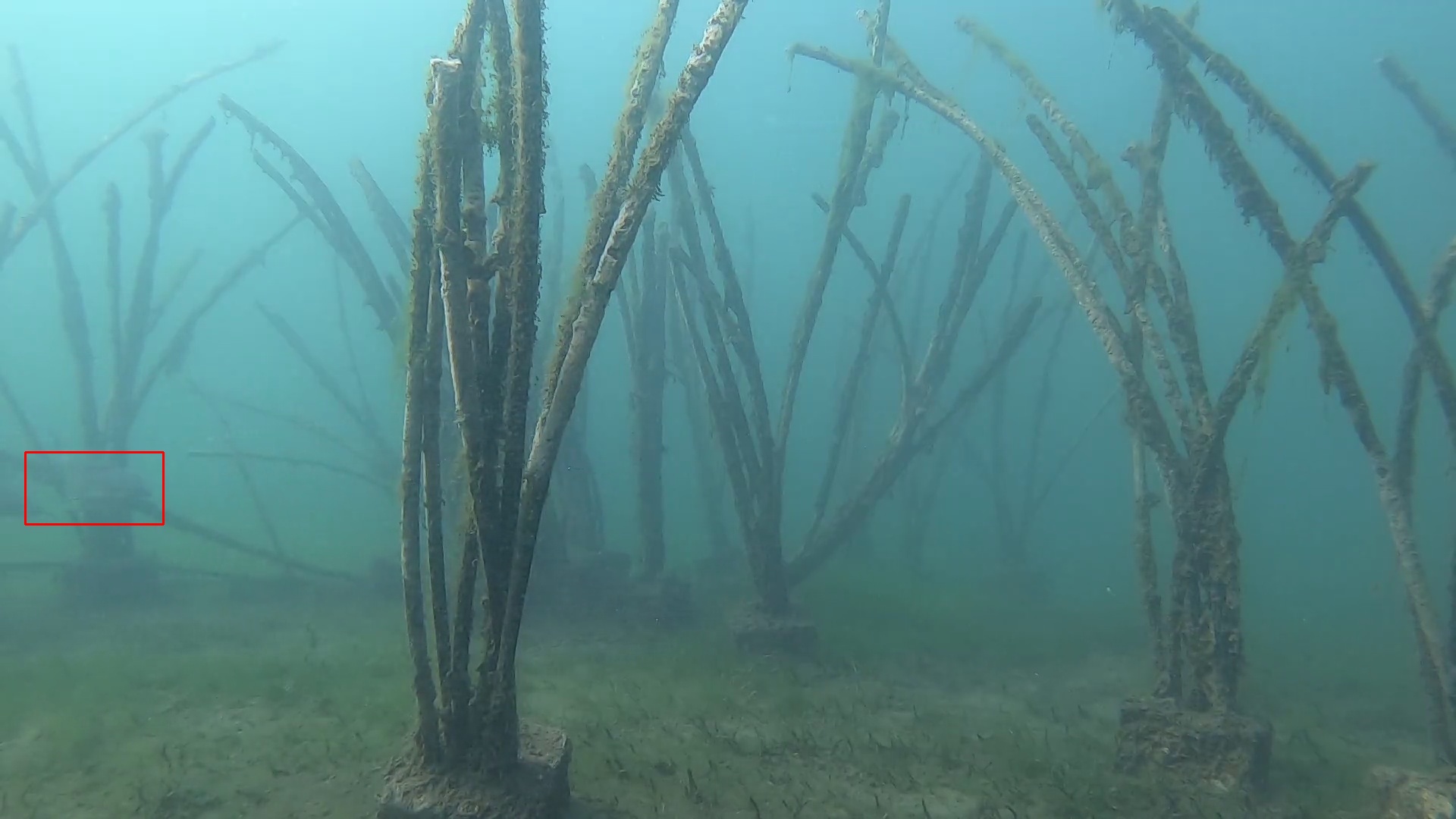} &
    & 
    \includegraphics[width=\linewidth]{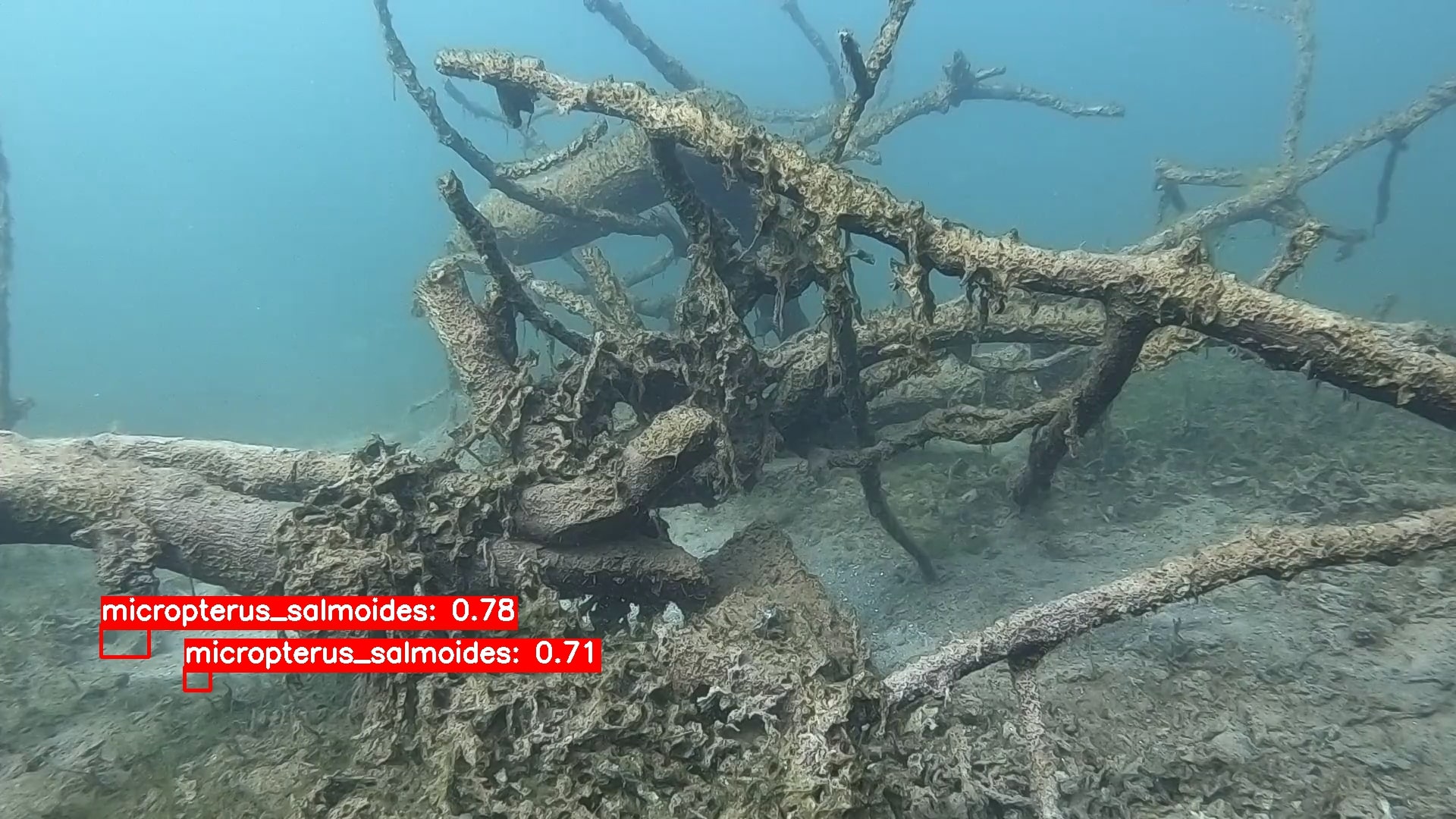} &
    \includegraphics[width=\linewidth]{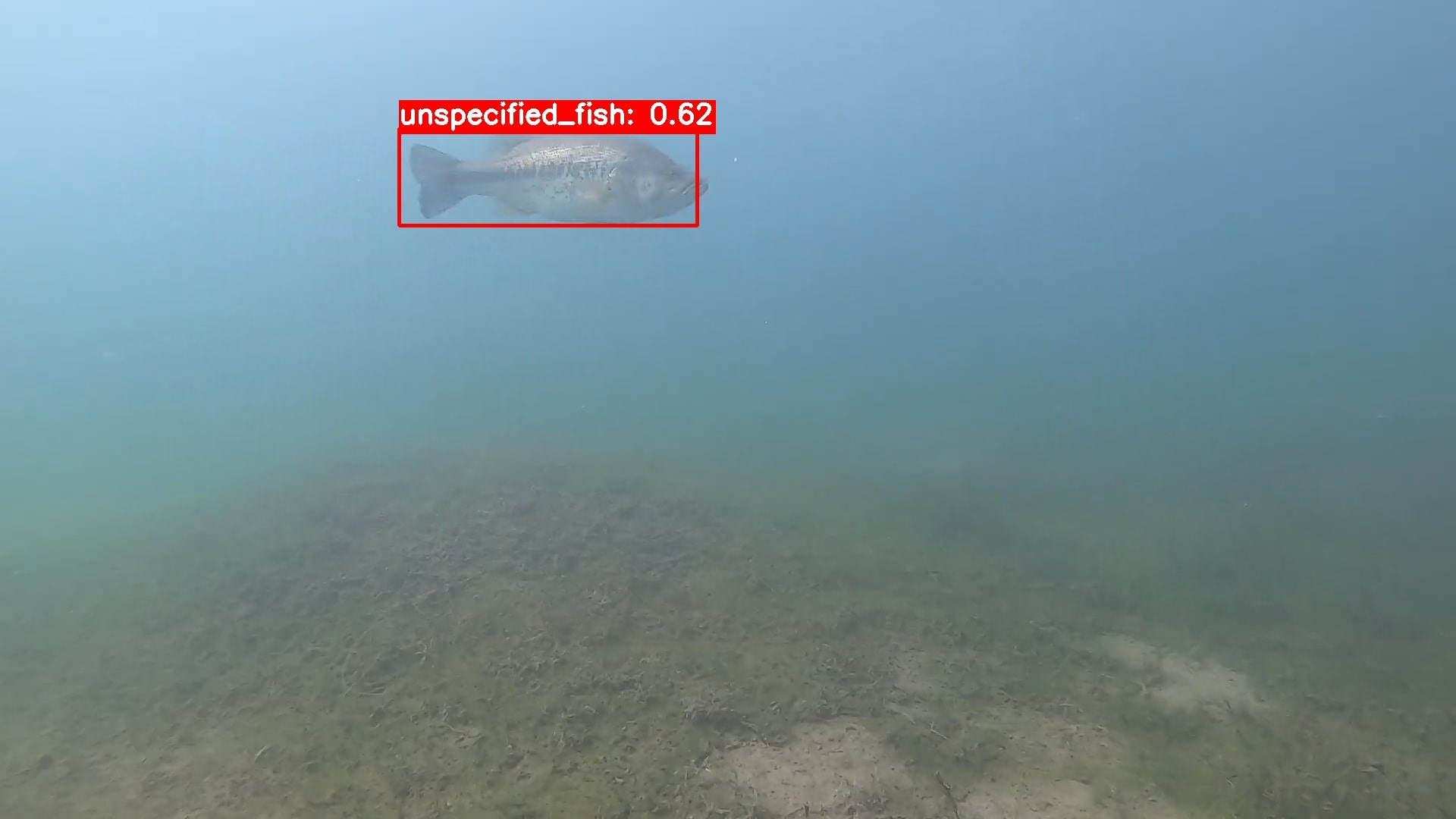} &
    \includegraphics[width=\linewidth]{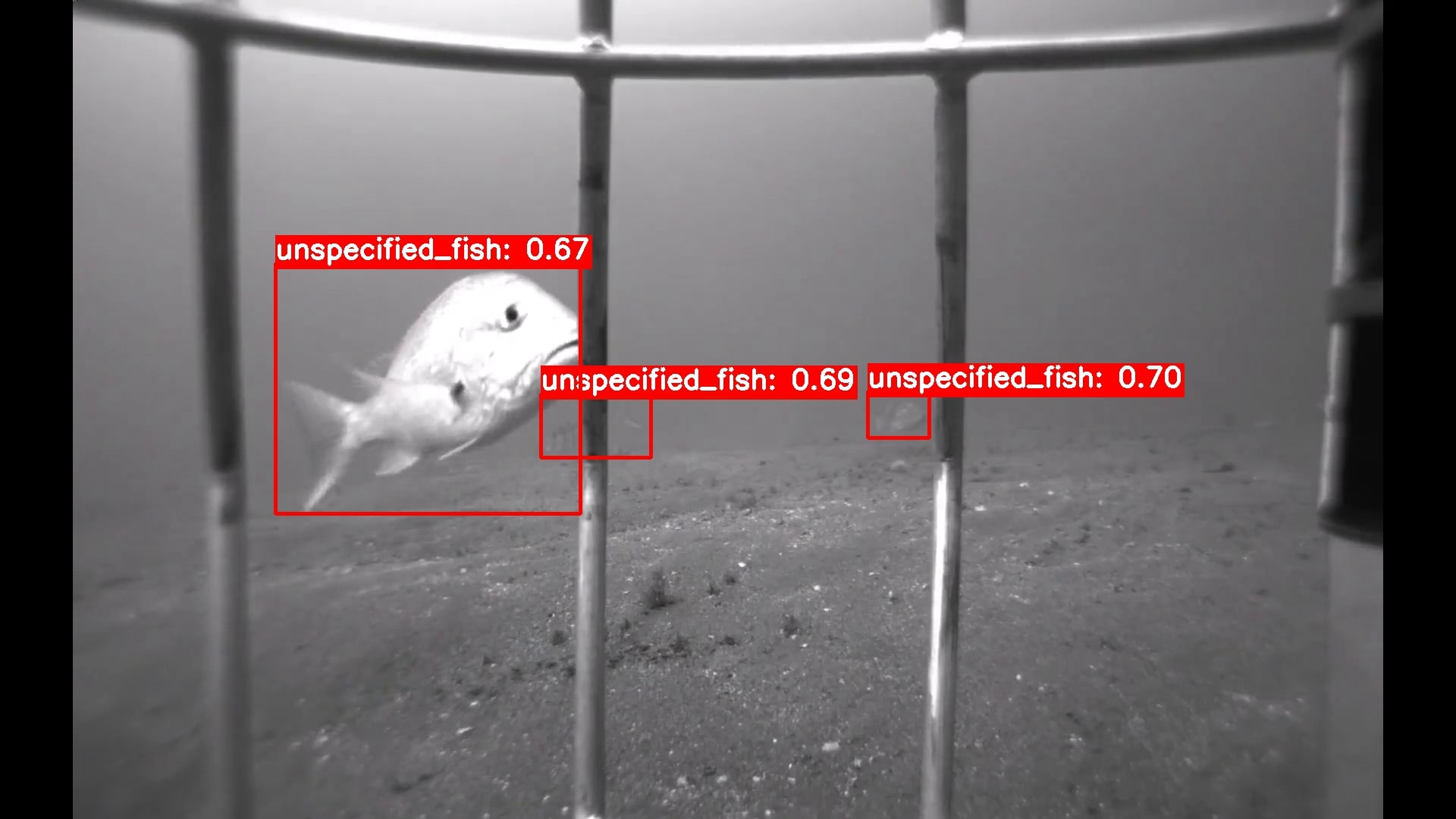} \\[-4pt]
    
    \scriptsize FeatEnHancer (YOLO) \cite{hashmi2023featenhancer} &
    \includegraphics[width=\linewidth]{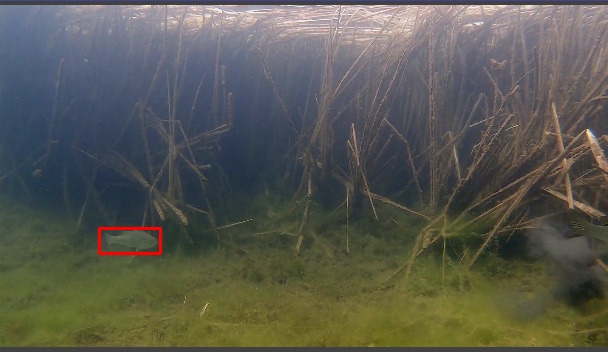} &
    \includegraphics[width=\linewidth]{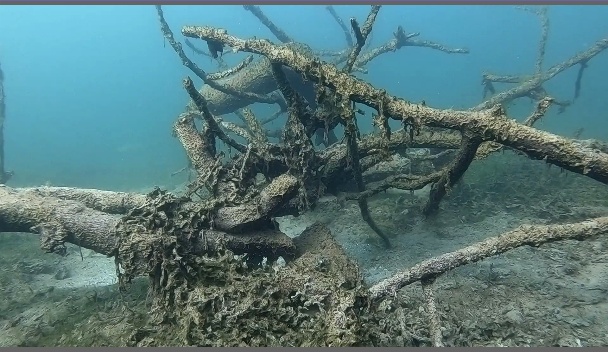} &
    \includegraphics[width=\linewidth]{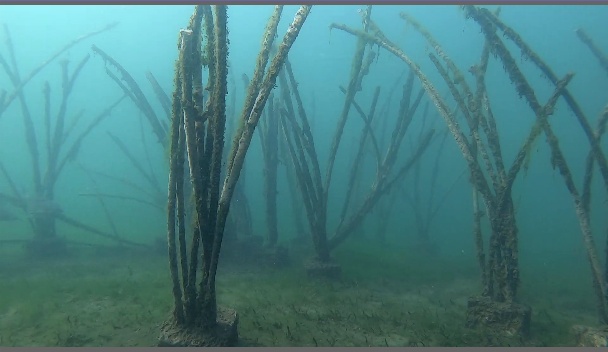} &
    & 
    \includegraphics[width=\linewidth]{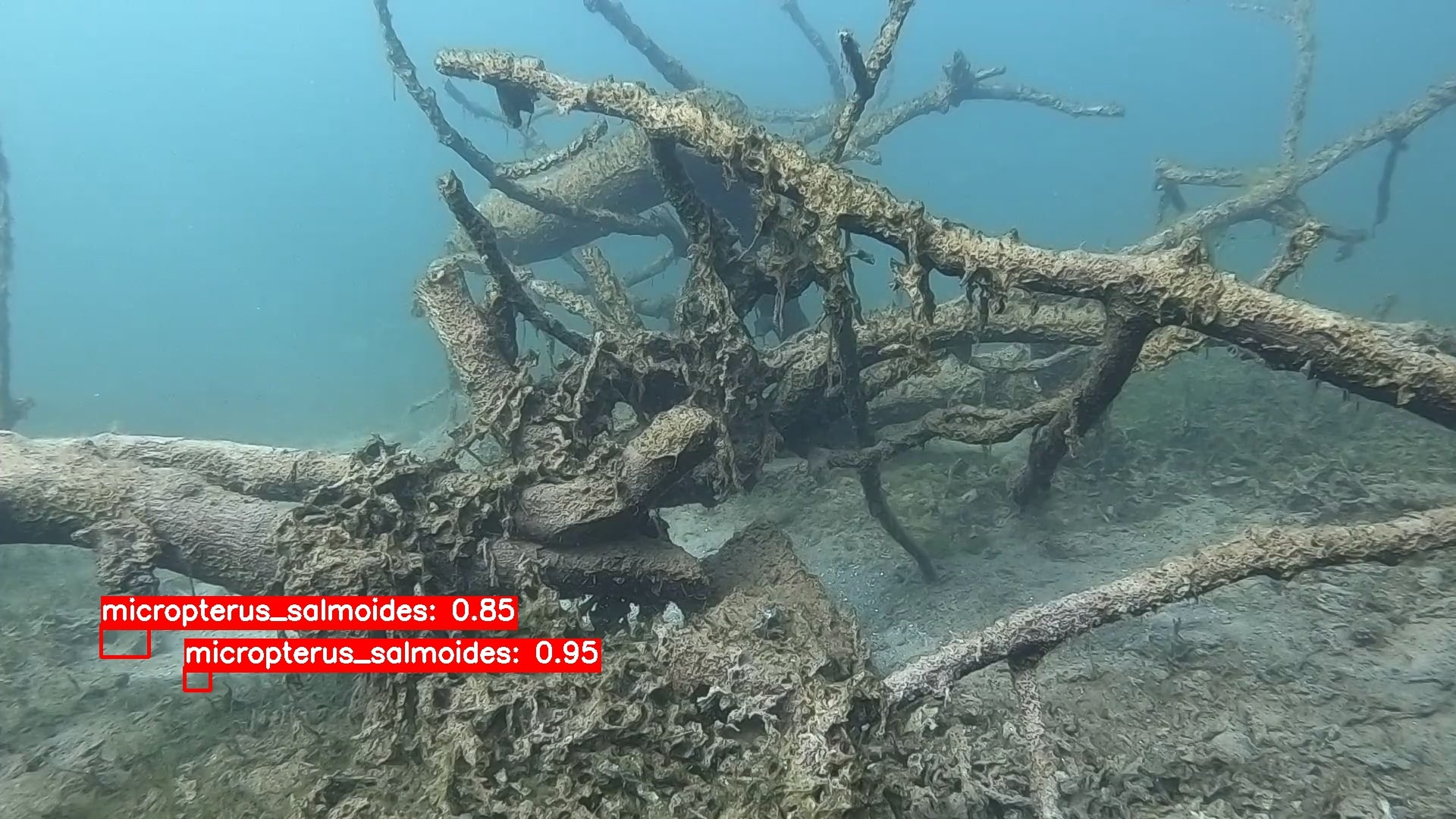} &
    \includegraphics[width=\linewidth]{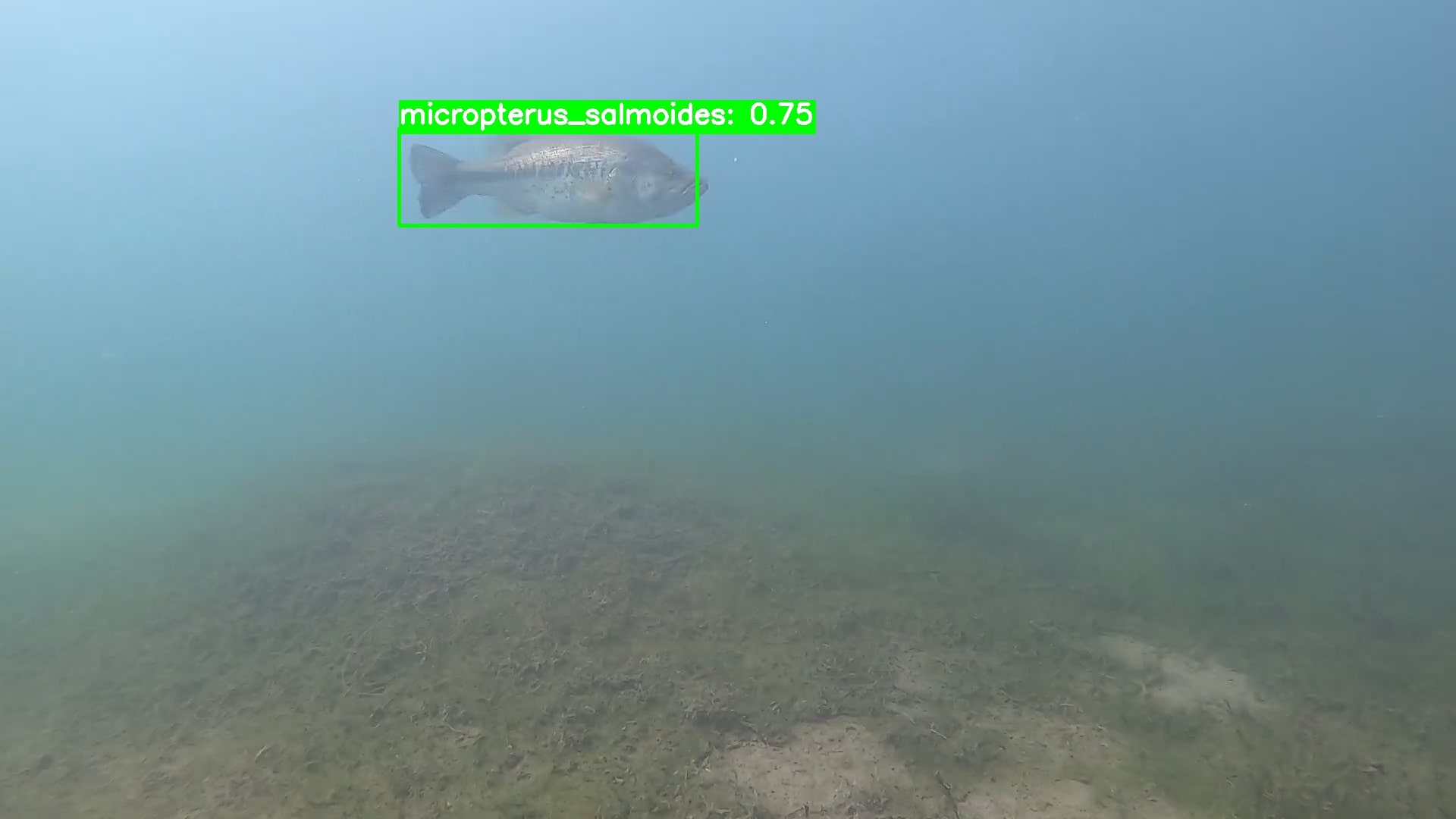} &
    \includegraphics[width=\linewidth]{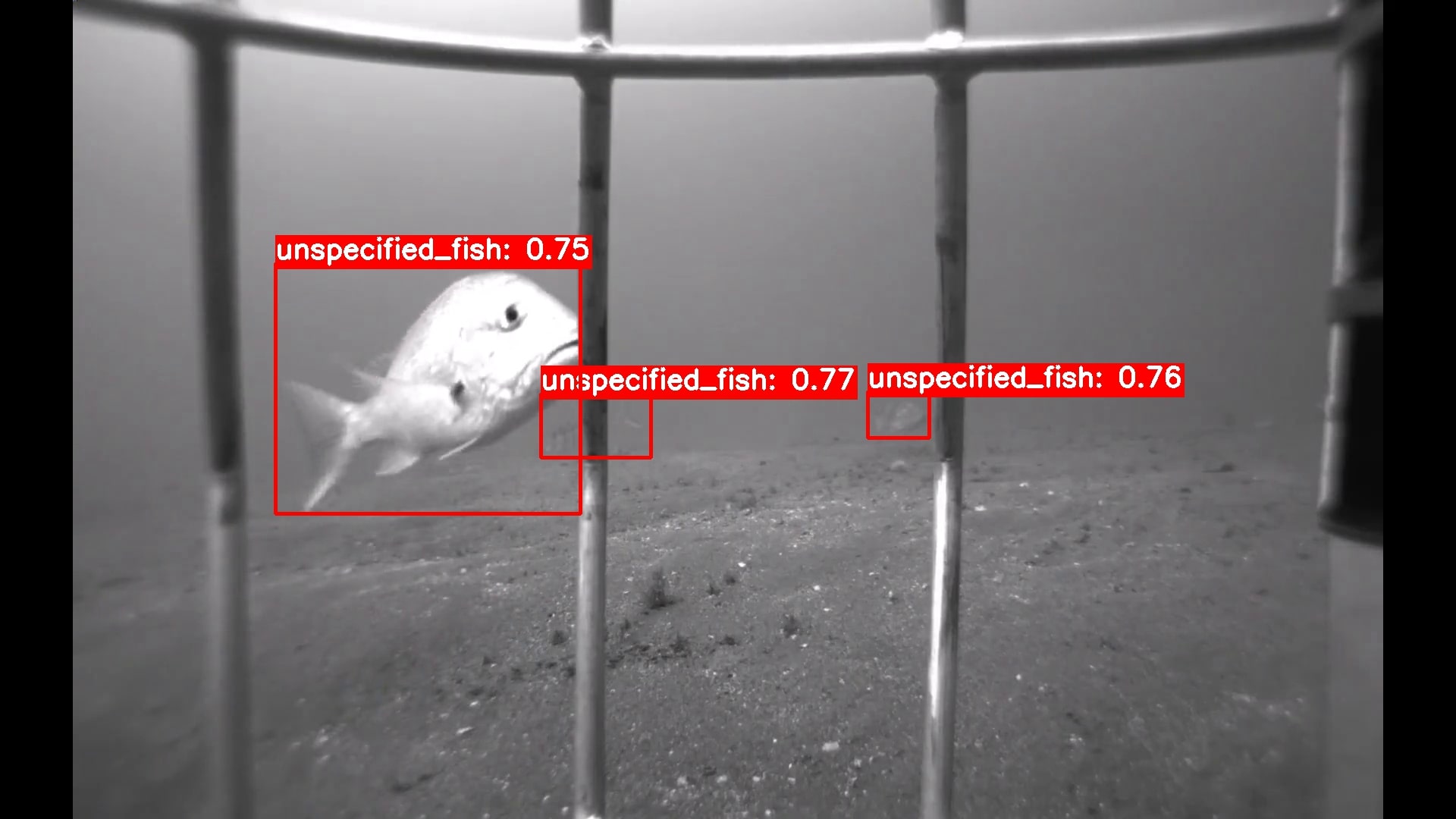} \\[-4pt]
    
    \scriptsize AquaFeat (YOLO) \cite{Silva2025AquaFeat_unpub} &
    \includegraphics[width=\linewidth]{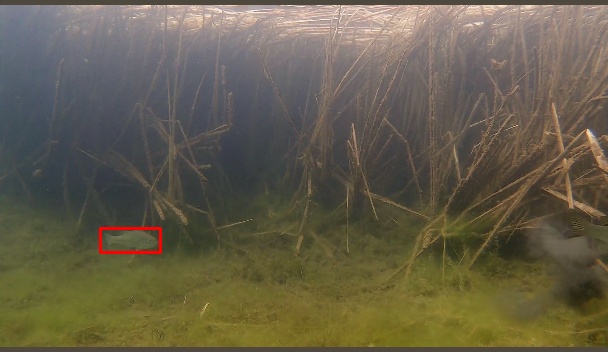} &
    \includegraphics[width=\linewidth]{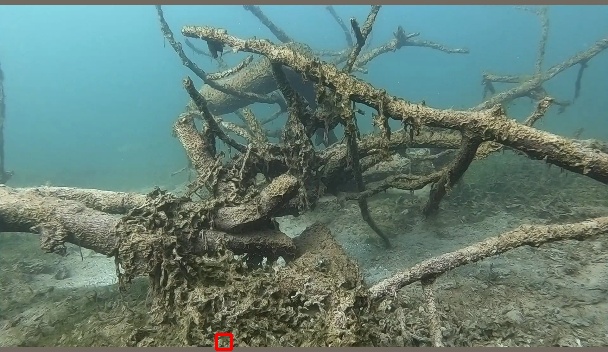} &
    \includegraphics[width=\linewidth]{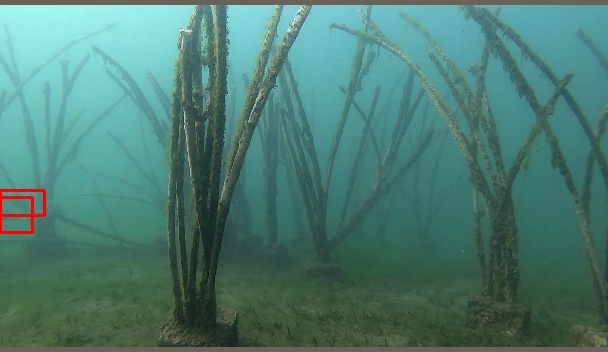} &
    & 
    \includegraphics[width=\linewidth]{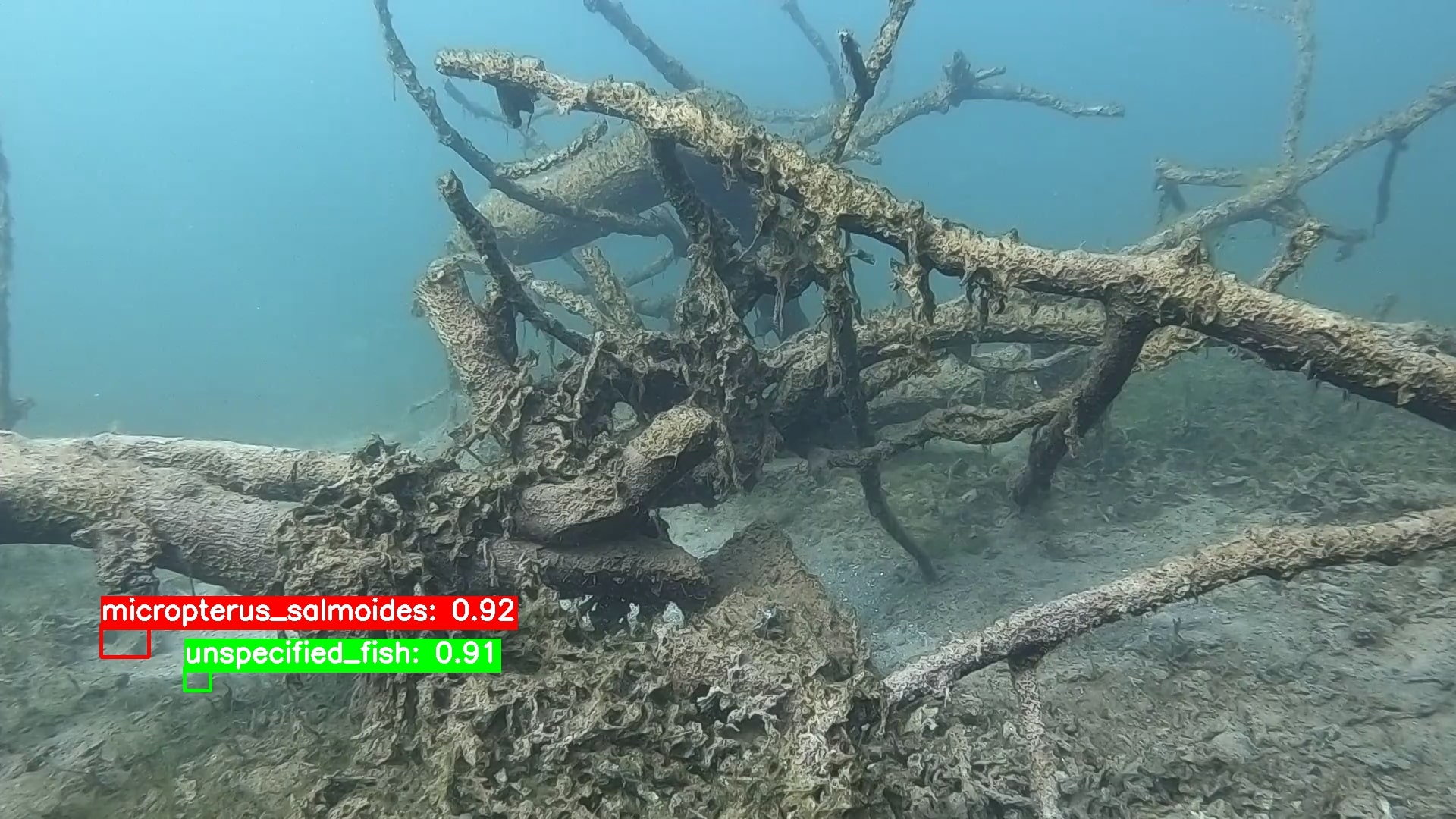} &
    \includegraphics[width=\linewidth]{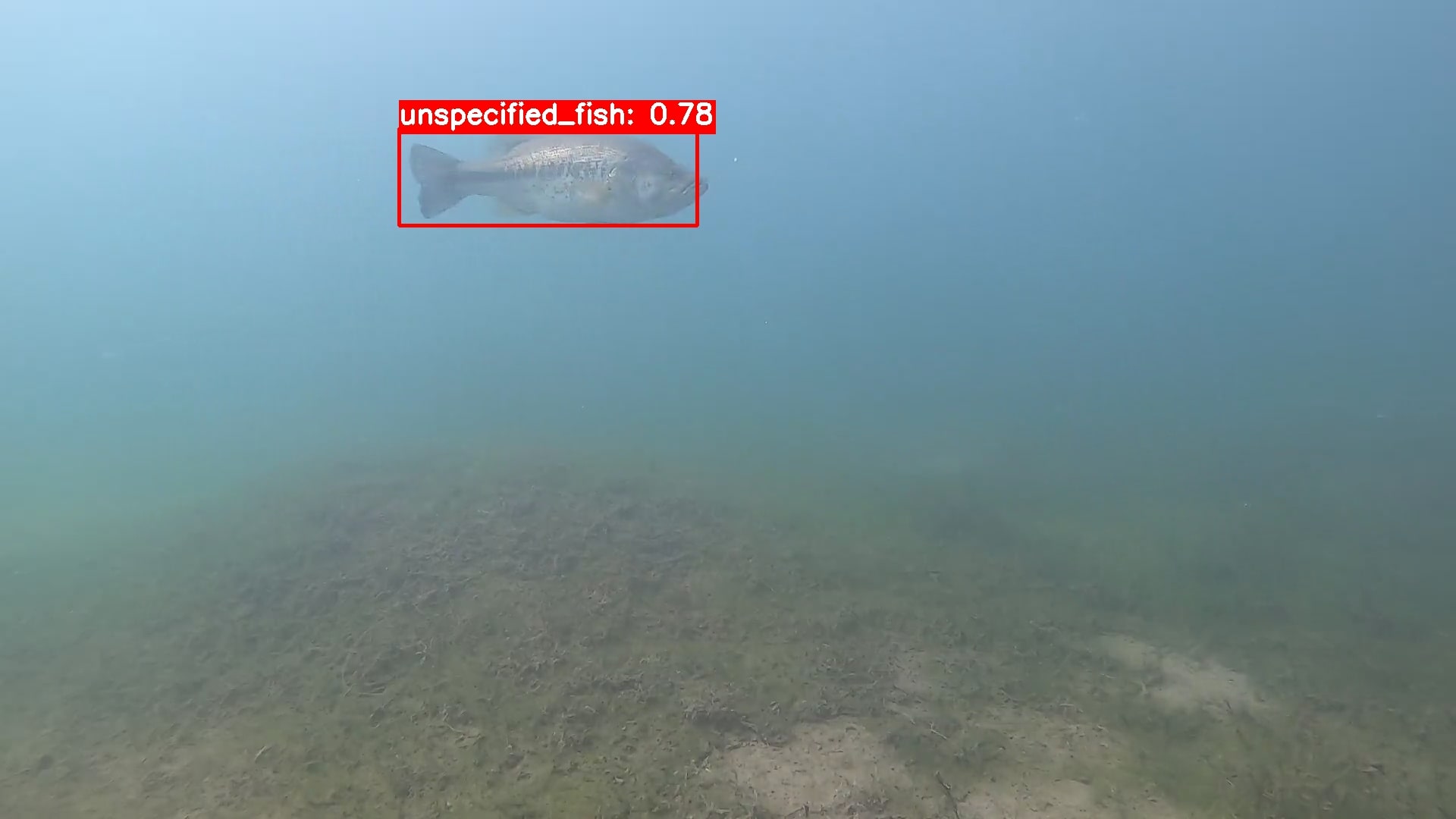} &
    \includegraphics[width=\linewidth]{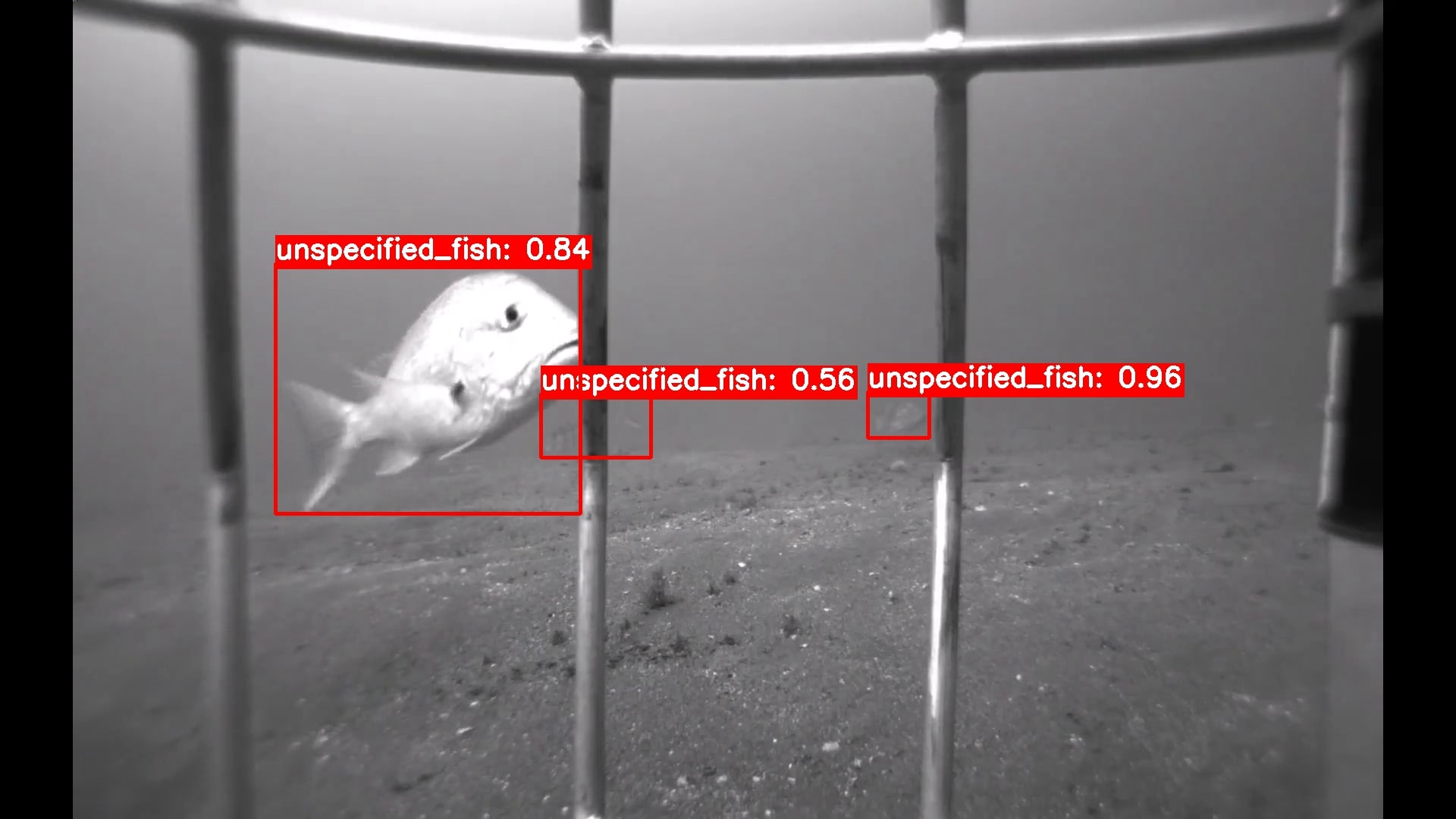} \\[-4pt]
    
    \scriptsize AquaFeat+ (Ours) (YOLO) &
    \includegraphics[width=\linewidth]{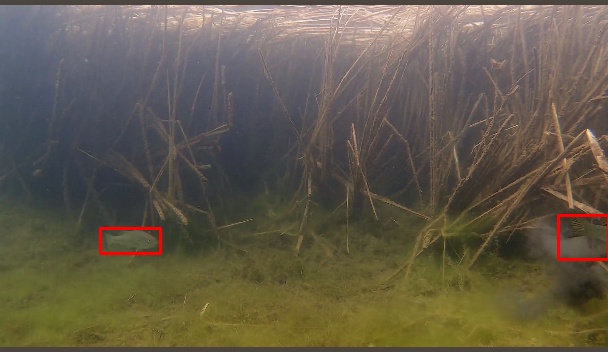} &
    \includegraphics[width=\linewidth]{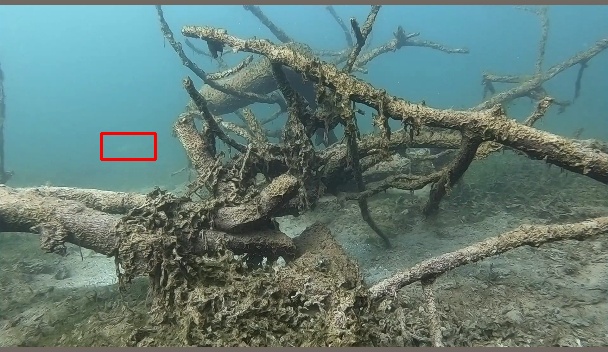} &
    \includegraphics[width=\linewidth]{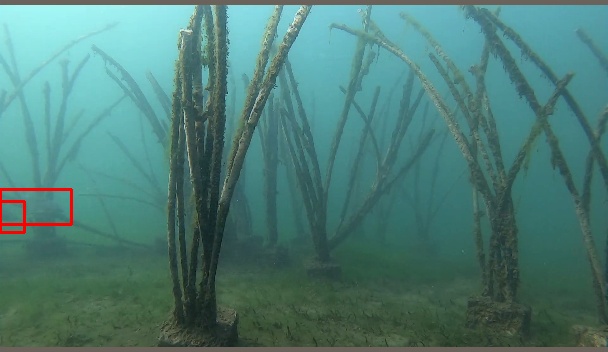} &
    & 
    \includegraphics[width=\linewidth]{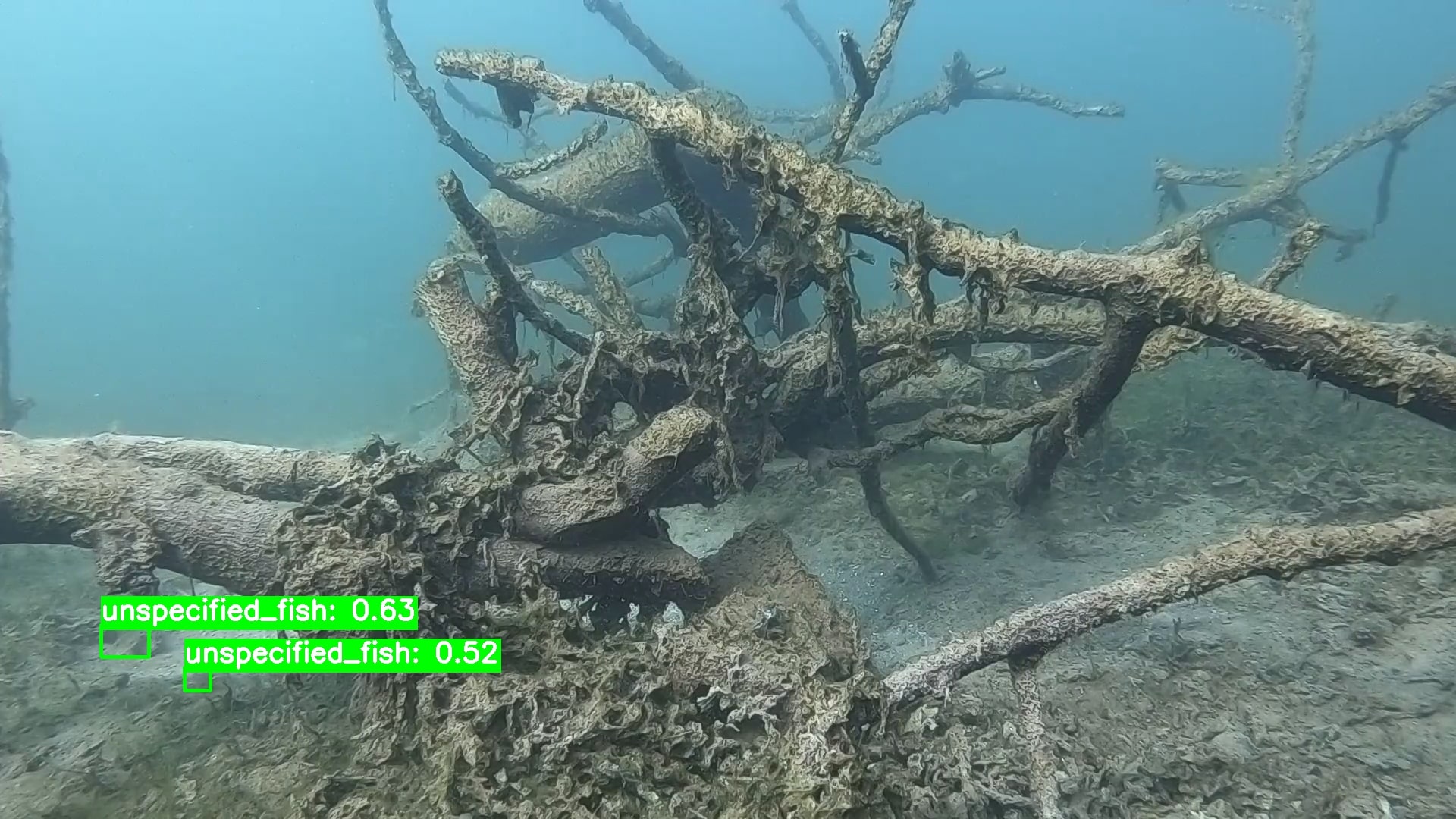} &
    \includegraphics[width=\linewidth]{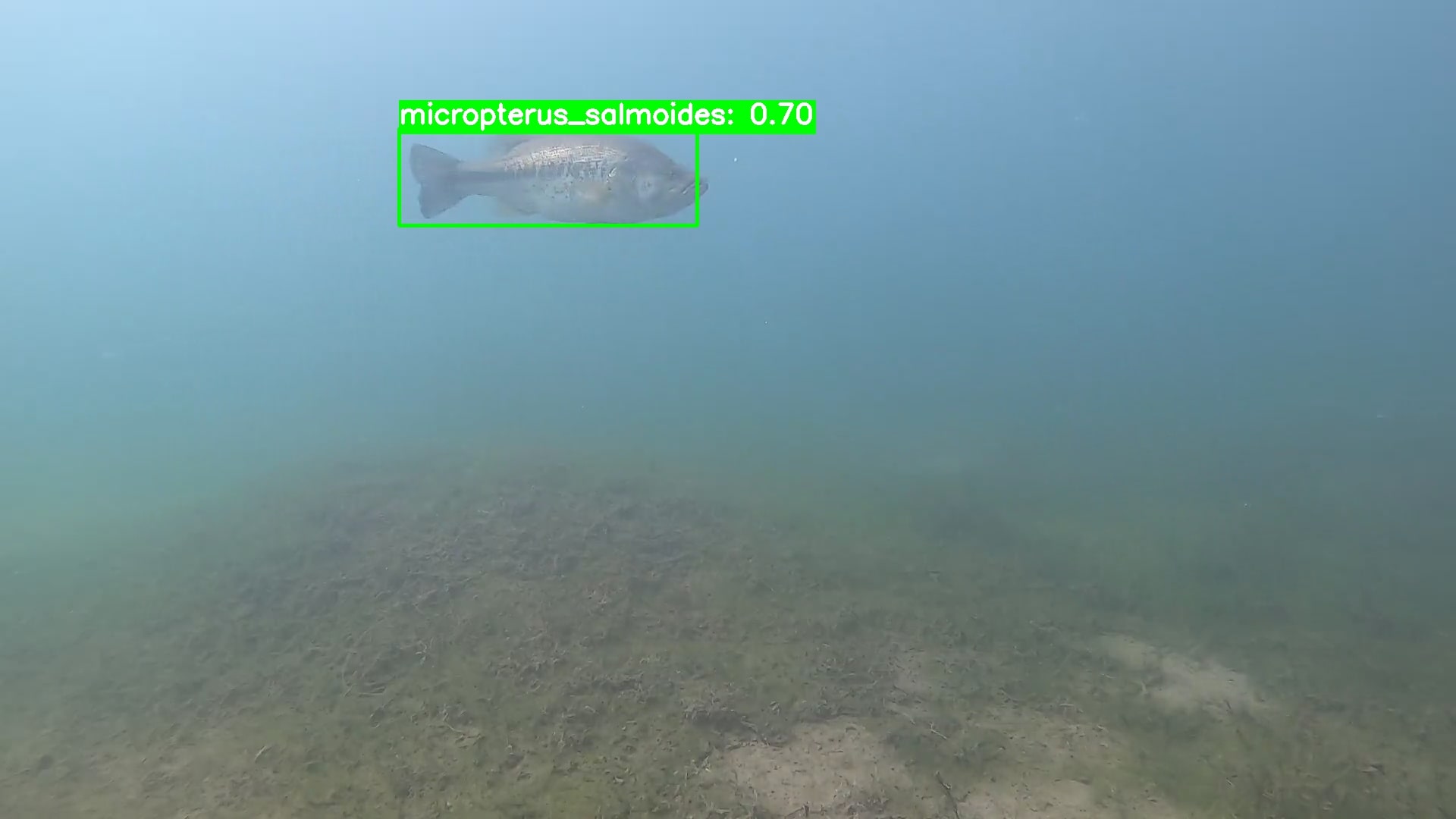} &
    \includegraphics[width=\linewidth]{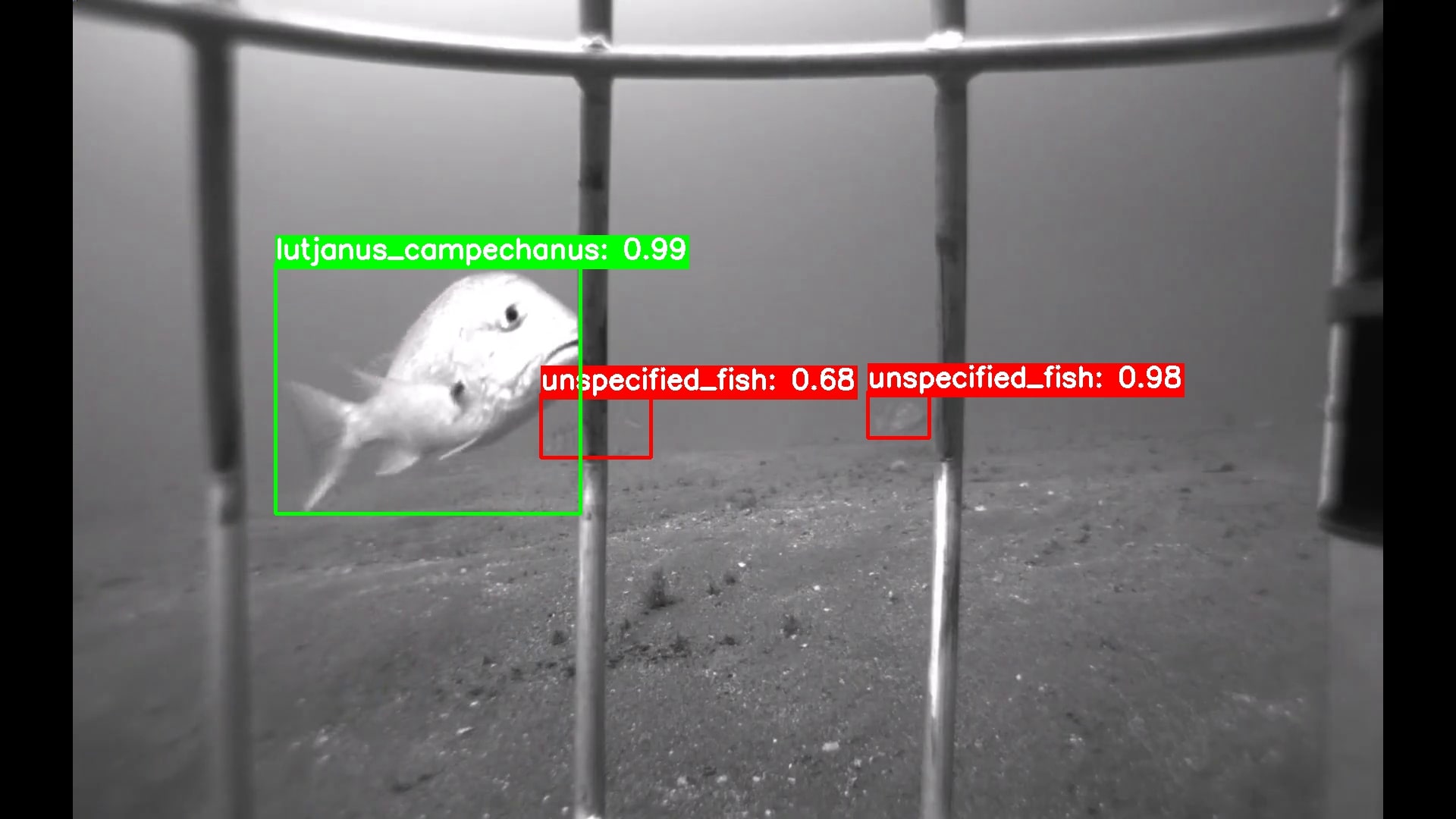} \\[4pt]
    
    & \multicolumn{3}{c}{\scriptsize \textbf{(a) Object Detection Results}} & & \multicolumn{3}{c}{\scriptsize \textbf{(b) Classification Results}} \\
    \end{tabular}}
    \caption{Qualitative comparison of model performance on the FishTrack23 dataset. (a) Bounding box predictions for the object detection task (YOLOv8m). (b) Species identification results for the classification task (YOLOv11s-cls). For the classification results, blue bounding boxes denote the ground truth, green boxes indicate a correct classification by the model, and red boxes signify a misclassification.}
    \label{fig:qualitative}
\end{figure*}

In qualitative analysis, we compared the object detection and classification results of our proposed AquaFeat+ enhancement method with those of other visual enhancement techniques working along with YOLOv8m and the baseline YOLOv8m model. As illustrated in Fig. \ref{fig:qualitative} (a), AquaFeat+ improved detection by enhancing image visibility and enabling more precise object localization. Most notably, as demonstrated in the scenario in the first column, AquaFeat+ was the only method capable of correctly detecting a partially occluded object at the image border. In the second column, only our method correctly detected the fish. In the third column, both AquaFeat models correctly identified both fish in the scene, while FeatEnHancer found neither, and neither did the baseline model.

The classification results are presented in Fig. \ref{fig:qualitative} (b). It can be observed that AquaFeat+ outperformed all other approaches. In the first column, our method correctly classified both fish within the image, whereas the original AquaFeat identified only one. In the second column, only our method and FeatEnhancer achieved a correct classification. Finally, in the last column, AquaFeat+ was the only method capable of correctly identifying one of the fish in the images. 

\begin{figure}
\centering 
\includegraphics[width=\columnwidth]{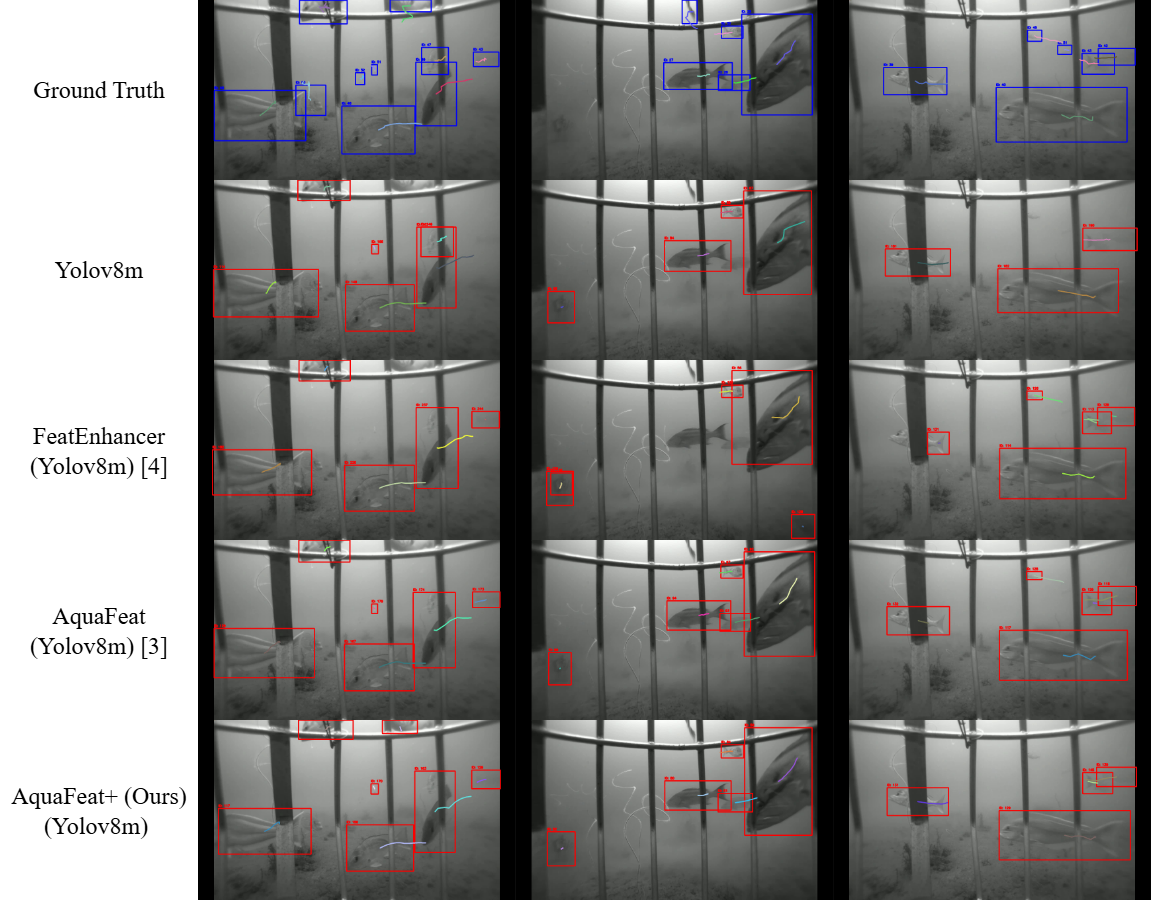}
\caption{Qualitative results of model performance across different methods in tracking in the FishTrack23 dataset. Blue boxes indicate the Ground truth, and red boxes the objects detected, while the colored lines inside them represent the movement in the ground truth and are verified by the methods.}
\label{fig:quali_track}
\end{figure}

Ultimately, the tracking evaluation is shown in Fig. \ref{fig:quali_track}. These results highlight the detector's critical impact on maintaining consistent IDs and accurate trajectories, particularly during occlusions. For instance, the AquaFeat models demonstrate a superior ability to handle occluded objects, with the Plus variant being the most effective. A key example in this sequence is the detection of fish behind bars, a feat accomplished only by our AquaFeat models.

\subsection{Quantitative Evaluation}

This section presents a quantitative analysis of our proposed method, AquaFeat+, comparing its performance against baseline and state-of-the-art methods across object detection, classification, and tracking tasks. The evaluations were performed on an underwater video dataset, and the detailed results are summarized in Tables \ref{detecção}, \ref{classification}, and \ref{tracking}.

\subsubsection{\textbf{Object Detection Metrics}} Our evaluation employed standard computer vision metrics. We utilized Precision, the fraction of correct detections among all predictions, and Recall, the fraction of all relevant objects that were successfully detected \cite{manning2009introduction}. We also report the mean Average Precision (mAP) under two standard settings: using an Intersection over Union (IoU) threshold of 50\% ($mAP_{0.5}$), and averaged over IoU thresholds from 0.50 to 0.95 ($mAP_{0.5:0.95}$). Detailed results are presented in Table \ref{detecção}.

\begin{table}[h]
\caption{Metrics Results for Object Detection}
\label{detecção}
 \begin{center}
 \resizebox{\columnwidth}{!}{%
 \begin{tabular}{|c|c|c|c|c|c|c|}
 \hline
 \textbf{Method} & \textbf{Precision$\uparrow$} & \textbf{Recall$\uparrow$} & \textbf{F1-Score$\uparrow$} & \textbf{mAP50$\uparrow$} & \textbf{mAP50-95$\uparrow$}\\
 \hline
 YOLOv8m & \textbf{0.792} & 0.582 & 0.677 & 0.528 & 0.319 \\
 \hline
 YOLOv10s & \underline{0.787} & 0.542 & 0.642 & 0.486  & 0.277 \\
  \hline
 FeatEnHancer \cite{hashmi2023featenhancer} (YOLOv8m) & 0.753 & 0.582 & 0.657 & 0.515  & 0.293 \\ 
 \hline 
 FeatEnHancer \cite{hashmi2023featenhancer} (YOLOv10s) & 0.732 & 0.590 & 0.654 & 0.503 & 0.293 \\
 \hline
 AquaFeat (YOLOv8m) \cite{Silva2025AquaFeat_unpub} & 0.746 & \underline{0.624} & 0.680 & 0.554 & 0.332 \\ 
  \hline
  AquaFeat (YOLOv10s) \cite{Silva2025AquaFeat_unpub} & 0.754 & \textbf{0.625} & 0.684 & \textbf{0.557} & \textbf{0.336} \\
 \hline
 \textbf{AquaFeat+} (YOLOv8m) & 0.767 & \underline{0.624} & \textbf{0.688} & \underline{0.556}  & 0.332 \\
 \hline
 \textbf{AquaFeat+} (YOLOv10s) & 0.770 & 0.619 & \underline{0.687} & 0.553 & \underline{0.333} \\
  \hline
 \end{tabular}
 }
 \end{center}
 \end{table}

The data analysis reveals that AquaFeat+ (YOLOv8m) achieved the highest F1-Score $0.688$, indicating the best balance between precision and recall among all evaluated methods. Although the AquaFeat (YOLOv10s) version recorded the highest values for $mAP_{0.5}$ $0.557$ and $mAP_{0.5:0.95}$ $0.336$, the performance of AquaFeat+ remained extremely competitive, with values of $0.556$ and $0.332$, respectively.

Notably, while the baseline YOLOv8m achieved the highest precision $0.792$, its recall was significantly lower $0.582$. In contrast, the AquaFeat and AquaFeat+ methods demonstrated a superior ability to correctly identify the objects of interest, reaching recall values of up to $0.625$, which justifies their higher F1-Scores and mAP. These results demonstrate that our approach effectively improves the model's overall detection capability, rather than merely optimizing precision at the expense of detecting relevant objects.

\subsubsection{\textbf{Classification Metrics}} The evaluation of the proposed approach was conducted using widely adopted metrics in computer vision, namely precision, recall, F1-score, and accuracy, which collectively offer a robust and comprehensive assessment of model performance.

Table~\ref{classification} presents the comparative results across different classification methods. FeatEnHancer, Aquafeat, and AquaFeat+, which are methods that work alongside others, were paired with the YOLOv11s-cls model. Overall, AquaFeat+ achieved the best performance, surpassing existing approaches in three out of four metrics. Obtaining the highest precision $0.816$, recall $0.791$, and F1-score $0.791$, showing a consistent ability to identify and balance positive predictions correctly. Although ConvNeXt reached the best accuracy $0.862$, its recall and F1-score were notably lower, indicating limitations in detecting positive samples despite overall correctness. Compared to AquaFeat, our enhanced AquaFeat+ model demonstrated clear improvements, confirming the effectiveness of the proposed enhancements for robust fish classification. 

\begin{table}[h]
\caption{Metrics Results for Classification}
\label{classification}
 \begin{center}
 \resizebox{\columnwidth}{!}{%
 \begin{tabular}{|c|c|c|c|c|}
 \hline
 \textbf{Method} & \textbf{Precision$\uparrow$} & \textbf{Recall$\uparrow$} & \textbf{Accuracy$\uparrow$} & \textbf{F1-Score$\uparrow$}\\
 \hline
 YOLOv11s-cls & 0.723 & 0.764 & 0.764 & 0.737 \\
 \hline
 FeatEnHancer \cite{hashmi2023featenhancer} & 0.746 & \underline{0.779} & 0.779 & 0.752 \\ 
 \hline
 ResNeXt \cite{xie2017aggregated} & $0.681$ & $0.605$ & \underline{0.834} & $0.618$ \\
 \hline
 ConvNeXt \cite{liu2022convnet} & $0.716$ & $0.619$ & $\mathbf{0.862}$ & $0.646$ \\
 \hline
 WildFish \cite{zhuang2018wildfish} & $0.642$ & $0.448$ & $0.773$ & $0.440$ \\
 \hline
 AquaFeat \cite{Silva2025AquaFeat_unpub} & \underline{0.798} & 0.765 & 0.765 & \underline{0.766} \\ 
 \hline
 AquaFeat+ \textbf{(Ours)}  & \textbf{0.816} & \textbf{0.791} & 0.791 & \textbf{0.791} \\
 \hline
 \end{tabular}
 }
 \end{center}
 \end{table}


\subsubsection{\textbf{Tracking Metrics}}
We quantitatively evaluate our tracker using metrics from the TrackEval toolkit \cite{luiten2020trackeval}, with ByteTrack \cite{zhang2022bytetrack} as our baseline. Key metrics include MOTA (Multi-Object Tracking Accuracy), which summarizes overall errors but is sensitive to detector performance; HOTA (Higher Order Tracking Accuracy), a more balanced metric that separates detection (DetA) from association (AssA) accuracy \cite{luiten2021hota}; and IDF1, which measures long-term identity consistency.

Our results show that AquaFeat models consistently rank in the top two across most metrics. Notably, our best model, AquaFeat+ (YOLOv10s), achieves the highest score in the key HOTA metric, followed closely by the original AquaFeat (YOLOv8m). This indicates a superior balance between detection and association performance. The effectiveness of the association is further confirmed by AquaFeat+ achieving the top AssA score, indicating a significant improvement in continuous identification. An exception is the DetA metric, where the baseline model performs best; however, the classic AquaFeat model secures the second-best score. While leading in HOTA, AquaFeat+ shows a higher number of false positives. Despite this, it outperforms other state-of-the-art models and also achieves the second-best IDF1 score.

\begin{table}[h]
\caption{Metrics Results for Tracking}
\label{tracking}
 \begin{center}
 \resizebox{\columnwidth}{!}{%
 \begin{tabular}{|c|c|c|c|c|c|c|}
 \hline 
 \textbf{Method} & \textbf{HOTA$\uparrow$} & \textbf{MOTA$\uparrow$} & \textbf{DetA$\uparrow$} & \textbf{AssA$\uparrow$} & \textbf{IDF1$\uparrow$}\\
 \hline
 YOLOv8m & 52.75 & 53.782 &\textbf{51.42}& 54.408 &65.099 \\
 \hline
 YOLOv10s & 49.153 &51.802 & 47.328& 51.467   & 64.221 \\
 \hline
 FeatEnHancer \cite{hashmi2023featenhancer} (YOLOv8m) & 47.475 & 37.228 & 41.147& 54.969   & 59.42\\ 
 \hline 
 FeatEnHancer \cite{hashmi2023featenhancer} (YOLOv10s) & 49.132 & 44.198 &45.114 & 53.681 &60.843 \\
 \hline
 AquaFeat (YOLOv8m) \cite{Silva2025AquaFeat_unpub} & \underline{54.715} & \textbf{55.762} & \underline{50.931} & \underline{59.144} &\textbf{ 68.414} \\ 
  \hline
  AquaFeat (YOLOv10s) \cite{Silva2025AquaFeat_unpub} & 52.143 & \underline{55.208} & 50.485 & 54.096&63.358\\
 \hline
 \textbf{AquaFeat+} (YOLOv8m) & 54.197 & 54.970  & 50.105 & 58.904  & 67.631 \\
 \hline
 \textbf{AquaFeat+} (YOLOv10s) & \textbf{55.206} &55.01 & 50.896 &\textbf{ 60.194 } & \underline{68.094} \\
  \hline
 \end{tabular}
 }
 \end{center}
 \end{table}

\section{CONCLUSIONS}

This work presented AquaFeat+, a model that enhances hierarchical features to optimize object detection, classification, and tracking tasks in underwater videos. The experimental results demonstrated that the method can improve task-level applications, such as object detection, classification, and tracking, making scene analysis more reliable for robotic decision-making. The flexibility of AquaFeat+ was demonstrated in different YOLO architectures, reinforcing its potential for integration into diverse robotic platforms. Future work will involve extending AquaFeat's application to more tasks, such as depth estimation and semantic segmentation. Additionally, we are planning to create new real datasets and evaluate the AquaFeat+ in other state-of-the-art learning architectures.

\addtolength{\textheight}{-12cm}   









\bibliographystyle{IEEEtran}
\bibliography{referencias} 

\end{document}